\pdfoutput=1
\documentclass[journal]{IEEEtran}
\usepackage{cite}
\usepackage{amsmath,amssymb,amsfonts}
\usepackage{graphicx}
\usepackage{algorithm}
\usepackage{algorithmic}
\usepackage{array}
\usepackage{tabularx}
\usepackage{multirow}
\usepackage{textcomp}
\usepackage{bm}
\usepackage{url}
\usepackage[english]{babel}
\usepackage[colorlinks,citecolor=blue,linkcolor=blue,urlcolor=blue]{hyperref}
\hypersetup{
pdftitle={Quantum Versus Classical Machine Learning for Neuropsychiatric Disorders: A Feature-Matched Multi-Cohort Evaluation},
pdfauthor={Utsav Poudel, Anuj Nepal, Vyas Viswanath, Subramaniyaswamy Vairavasundaram},
pdfkeywords={quantum machine learning, functional connectivity, fMRI, neuropsychiatric disorders, quantum kernel methods}}
\newcommand{\orcid}[1]{\href{https://orcid.org/#1}{\raisebox{-0.12ex}{\includegraphics[height=1.7ex]{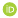}}}}

\begin{document}

\title{Quantum Versus Classical Machine Learning for Neuropsychiatric Disorders: A Feature-Matched Multi-Cohort Evaluation}

\author{Utsav~Poudel\,\orcid{0000-0002-5800-1263},
Anuj~Nepal\,\orcid{0009-0009-8177-3278},
Vyas~Viswanath,
and~Subramaniyaswamy~Vairavasundaram\,\orcid{0000-0001-5328-7672}%
\thanks{This work has been submitted to the IEEE for possible publication.
Copyright may be transferred without notice, after which this version may no
longer be accessible.}%
\thanks{This work was supported by Vellore Institute of Technology, Vellore, India.}%
\thanks{U.~Poudel, V.~Viswanath, and S.~Vairavasundaram are with the School of
Computer Science and Engineering, Vellore Institute of Technology, Vellore,
Tamil Nadu 632014, India (e-mail: utsav.poudel2022@vitstudent.ac.in;
vyas.viswanath2024@vitstudent.ac.in; subramaniyaswamy.v@vit.ac.in).}%
\thanks{A.~Nepal is with Universal Higher Education, Melbourne, Australia
(e-mail: anuj.n@uhe.edu.au).}%
\thanks{Corresponding author: Subramaniyaswamy Vairavasundaram.}}

\markboth{Preprint}%
{Poudel \MakeLowercase{\textit{et al.}}: Quantum Versus Classical Machine Learning for Neuropsychiatric Disorders}

\maketitle

\begin{abstract}
Resting-state functional connectivity is widely used in machine learning-based classification of neuropsychiatric disorders. Quantum machine learning models have been proposed for functional connectivity data, but few studies have evaluated them under leave-site-out validation or against matched classical models. We compared six quantum model families with classical ones using identical correlation-derived features across four public cohorts (4,308 subjects). Leave-site-out validation was used for the Autism Brain Imaging Data Exchange (ABIDE I), ADHD-200 (attention-deficit/hyperactivity disorder), and REST-meta-MDD (major depressive disorder) cohorts, while the UCLA Consortium for Neuropsychiatric Phenomics (UCLA-CNP) was evaluated using stratified 10-fold cross-validation. The quantum kernel was statistically indistinguishable from a matched classical kernel across all four cohorts, with the average accuracy difference within 1.4 percentage points at 95\% confidence. On ABIDE I, a hybrid quantum convolutional network achieved similar accuracy to its classical counterpart (0.645 vs. 0.641), while a trainable quantum kernel exceeded its classical counterpart by 0.064 in area under the curve (AUC). Quantum convolutional networks also achieved higher AUC than their matched classical networks on this cohort and showed smaller train--test gaps on all four cohorts, without an accuracy gain. Overall, most quantum models matched their classical counterparts, but none consistently outperformed them. This study, among the first quantum evaluations of ABIDE I under leave-site-out validation, provides a feature-matched benchmark of quantum and classical models comprising 284 matched comparisons. We further analyse fidelity-based graph construction and account for the observed quantum--classical parity through effective-dimension and eigenspectrum analyses. Code and per-fold results are available at \url{https://github.com/r11up/qml-fmri-baseline}.
\end{abstract}


\section{Introduction}\label{sec-intro}

Artificial intelligence is increasingly being applied to psychiatric diagnosis, prognosis, and decision support systems \cite{Poudel2025AIMentalHealth}, and resting-state functional magnetic resonance imaging (rs-fMRI) is widely used as one of the few tools that enable non-invasive analysis of intrinsic functional brain activity. Most rs-fMRI pipelines begin by building a functional connectivity graph whose edges encode statistical dependency between regions of interest (ROIs). The Pearson correlation coefficient remains one of the most widely used measures because of its computational efficiency and interpretability \cite{Vachha2025RestingState}. However, it captures only linear, pairwise relationships and may overlook non-linear or higher-order dependencies between brain regions. For conditions with subtle, heterogeneous presentation, such as autism spectrum disorder (ASD) \cite{Vidya2025AutismFMRI}, non-linear dependence may carry complementary information, as reported for schizophrenia \cite{motlaghian2022nonlinear}, but whether it improves generalisation to unseen acquisition sites remains an open question.

Graph neural networks (GNNs) address this limitation through a different mechanism. Rather than replacing the linear edge measure, they keep the correlation graph and recover non-linearity downstream, composing non-linear functions of a node's neighbourhood across successive message-passing layers. This paper investigates whether introducing non-linearity at the connectivity level, in the graph edges themselves, is advantageous compared with introducing it only within downstream models. The downstream route has a cost, since stacking layers runs into over-smoothing, where repeated aggregation over a dense, noisy graph pulls node representations toward a common mean \cite{jin2025oversmoothing}. The problem is worse when the topology is static and dense, which is the typical shape of a Pearson correlation matrix in fMRI.

A classical vector encoded into a quantum state maps into a Hilbert space where relations that were non-linear in the original coordinates can become close to linear \cite{Schuld2019FeatureHilbert}. We encode correlation-derived regional features as quantum states, use the fidelity between them to initialise the graph's topology, and train a classical edge kernel and the classifier on that support. The design is therefore a quantum-fidelity-initialised adaptive graph rather than an end-to-end quantum one.
The graph topology is determined by the encoding component of the circuit alone. Following standard connectivity pipelines, every model in this study receives features derived from the correlation matrix, so quantum and classical models see identical inputs and any difference between them reflects the similarity function rather than the input representation.

This design sits between two bodies of work that have, until recently, developed largely along separate tracks: classical adaptive graph learning, which treats the brain connectivity graph as a learnable object, and quantum machine learning applied to neuroimaging and other healthcare data, which has rarely been evaluated against matched classical baselines under leave-site-out validation. This work brings the two together and makes the following contributions:

\begin{itemize}

\item \textbf{A matched, multi-cohort benchmark}: a feature-matched comparison of six quantum model families with classical models across four public cohorts, using leave-site-out validation where cohort structure permits, among the first such evaluations to our knowledge.
\item \textbf{An analysis of fidelity-based graph construction}: a theoretical account of common-unitary cancellation and register-width-dependent concentration.
\item \textbf{A spectral comparison of quantum and classical kernels}: an effective-dimension, eigenspectrum and predictive comparison of quantum-fidelity and radial basis function (RBF) kernels under identical features.
\item \textbf{A systematic evaluation of quantum--classical differences}: $284$ matched comparisons across cohorts and metrics, with prediction-degeneracy screening and multiple-testing correction, supported by released code and per-fold results.

\end{itemize}

\section{Related Work}\label{sec-literature}

Machine learning has advanced quickly across psychiatry over the past decade \cite{Poudel2024AIpsychiatry}, and two lines of that work converge on this problem: classical adaptive graph learning, which treats the connectivity graph as a learnable object rather than a fixed correlation matrix, and quantum machine learning applied to neuroimaging and other biomedical data. Neither has yet produced a system that builds a functional-connectivity graph from quantum measurements and trains it against a clinical classification objective.

\subsection{Classical Adaptive Graph Learning for Functional Connectivity}
\label{sec:lit-classical}

The limits of static correlation graphs have pushed several groups toward making the graph itself a learnable parameter, optimised jointly with the network weights. AGMGC \cite{noman2025adaptive} replaces a predefined connectivity assumption with a learner module that infers structure from raw time series, augments it with a partial-correlation graph, and reaches $84.65\%$ accuracy on the Autism Brain Imaging Data Exchange (ABIDE), ahead of population-graph baselines. Its adjacency is still a softmax over classically learned projections: a more flexible classical similarity function, but a classical one.

A second trend folds in auxiliary modalities to refine the graph. DAGMNet \cite{Wang2025DAGMNet} fuses structural MRI (sMRI), fMRI, and phenotypic data (age, sex, and similar covariates) through a phenotype-guided population-graph module that prunes connections unsupported by clinical priors, reaching 91.59\% accuracy and a 96.80\% area under the receiver operating characteristic curve (AUC) on ABIDE~I. Section~\ref{sec:comparability} examines what that figure measures. Auxiliary context has also been paired with EEG, where a dual-tower model combining EEG with a graph-encoded environmental prior improved affective-state classification over EEG alone \cite{Poudel2026NeuroGeo}.

\subsection{Quantum Machine Learning in Neuroimaging and Healthcare}
\label{sec:lit-quantum}

A 2025 systematic review of quantum deep learning in neuroinformatics describes itself as a pioneering effort to compare quantum and classical approaches on brain data \cite{orka2025qdlneuroinformatics}, and a 2025 survey of quantum machine learning pairs the promise of hybrid models with the noise and qubit-scaling limits of current hardware \cite{lamichhane2025qmlsurvey}. What does exist clusters around a few entry points into brain data.

One entry point is the direct application of a quantum or hybrid circuit to a time series or an image. Choi et al.\ use a one-dimensional convolutional neural network (1-D CNN) front end with a quantum convolutional neural network (QCNN) on rs-fMRI \cite{choi2025quantumfmri}; Park et al.\ use a quantum time-series transformer on the same modality \cite{park2025quantumtransformer}. For ASD, Saranya and Menaka \cite{saranya2025quantumautism}, Bagoun et al.\ \cite{bagoun2024quantumasd} and Turjya and Fawakherji \cite{turjya2025p300asd} apply quantum classifiers to electroencephalography (EEG). None builds an inter-regional adjacency matrix.

The one quantum model we found evaluated on ABIDE~I functional connectivity is QG-WRN \cite{huang2026qgwrn}, which places a four-qubit parameterised circuit inside a graph convolutional residual network on the 116-region Automated Anatomical Labelling (AAL-116) atlas and reports $68.49\%$ accuracy on a random 80/20 split of $775$ subjects, a protocol its authors note does not decouple inter-site scanner variance.

\subsection{Quantum Architectures Beyond Static Fidelity}
\label{sec:beyond-fidelity}

The constraint this paper measures, that pairwise state fidelity concentrates as the register widens, is specific to kernels built from static pairwise overlaps, and a recent line of work avoids it by changing the quantum primitive rather than the register. The quantum time-series transformer of Park et al.~\cite{park2025quantumtransformer} realises attention through a Linear Combination of Unitaries and the Quantum Singular Value Transformation. Applied to resting-state fMRI, it processes temporal dependencies directly instead of collapsing them into a similarity matrix, operates at polylogarithmic complexity, and reports performance competitive with classical transformers on the Adolescent Brain Cognitive Development (ABCD) and UK~Biobank cohorts.

It targets the bottleneck that Section~\ref{sec:fidelity-collapse} measures without widening the register, and reports its gains as most pronounced in small samples, the regime in which our quantum arm overfits less (Section~\ref{sec:regularisation}) without gaining accuracy (Section~\ref{sec:lowdata}). We do not evaluate it here and identify it as the most natural next comparison for this benchmark.

\subsection{Positioning of This Work}
\label{sec:positioning}

Against the classical adaptive graph learning of Section~\ref{sec:lit-classical}, the difference is the similarity function: quantum fidelity in place of softmax attention or phenotype gating. Against the quantum-neuroimaging classifiers of Section~\ref{sec:lit-quantum}, the difference is architectural: an explicit adjacency, fidelity-initialised and classically weighted, feeding a graph-attention network rather than a per-region classifier. Against QG-WRN \cite{huang2026qgwrn}, the closest prior quantum model on the same cohort, the difference is protocol: whole acquisition sites are held out, the 200-region Craddock (CC200) parcellation is used, and every quantum model is set against a classical model matched on features, capacity, and training budget.

\subsection{Network-Level Findings Reported for These Cohorts}
\label{sec:networks}

Table~\ref{tab:cohorts} lists the functional networks the clinical literature reports as disrupted in each disorder, and Fig.~\ref{fig:networks} renders them on cortical and subcortical surfaces, each Yeo network \cite{yeo2011} coloured by its rank in that literature; the panel for the transdiagnostic UCLA Consortium for Neuropsychiatric Phenomics (UCLA-CNP) instead shows the CC200/Yeo parcellation, which doubles as the atlas reference throughout. For depression, the direction of default-mode change is contested, with the largest multi-site analysis reporting reduced connectivity in recurrent depression \cite{Yan2019RESTmetaMDD}. These are prior findings that situate the cohorts, not outputs of any model here; Section~\ref{sec:limitations} explains why models near chance cannot support network-level claims.

\section{Theoretical Framework}\label{sec-theory}

\subsection{Graph Representation of Functional Connectivity}

The human brain is modelled as a graph $G = (V, E)$ whose nodes $V = \{v_1, \dots, v_N\}$ are the $N$ brain regions defined by a brain parcellation, each carrying a feature vector $x_i \in \mathbb{R}^T$, the blood-oxygen-level-dependent (BOLD) signal time series over $T$ timepoints. The edges $E$ are the connections between regions, and the $N \times N$ adjacency matrix $A$ gives in $A_{ij}$ the strength of the functional connection between node $i$ and node $j$.

In classical functional connectivity analysis, $A$ is typically the Pearson correlation matrix:

\begin{equation}
A_{ij}^{\text{classical}} = \frac{\sum_{t=1}^T (x_{i,t} - \bar{x}_i)(x_{j,t} - \bar{x}_j)}{\sqrt{\sum_{t=1}^T (x_{i,t} - \bar{x}_i)^2} \sqrt{\sum_{t=1}^T (x_{j,t} - \bar{x}_j)^2}}
\end{equation}

Here $\bar{x}_i = \frac{1}{T}\sum_{t=1}^{T} x_{i,t}$ is the temporal mean of the signal of region $i$, so that $A_{ij}^{\text{classical}} \in [-1, 1]$.

Pearson correlation primarily captures linear, zero-lag, pairwise dependence, so it may not fully represent non-linear relationships, such as $x_j(t) = x_i(t)^2$, or higher-order relationships involving more than two regions at once.

\subsection{Quantum Feature Maps and Hilbert Spaces}
To compare regions through a non-linear similarity, we map each classical feature vector $x_i = [x_{i,1}, \dots, x_{i,d}]$ into a quantum state $|\psi_i\rangle$ in a Hilbert space $\mathcal{H}$ with a parameterised quantum circuit on $d$ qubits \cite{Havlicek2019QuantumFeature}. We use angle encoding, which needs one qubit and one rotation per feature, so the register grows linearly with feature dimension, and which avoids the normalisation and deep state-preparation circuits that amplitude encoding requires. Each feature $x_{i,k}$ becomes the angle of a rotation about the $Y$-axis on one qubit:
\begin{equation}
|\psi(x_i)\rangle = \bigotimes_{k=1}^{d} R_y(x_{i,k}) |0\rangle, \quad
R_y(\theta) = \begin{bmatrix} \cos\frac{\theta}{2} & -\sin\frac{\theta}{2} \\ \sin\frac{\theta}{2} & \cos\frac{\theta}{2} \end{bmatrix}.
\label{eq:angle}
\end{equation}
The amplitudes of the resulting state are products of $\cos(x_{i,k}/2)$ and $\sin(x_{i,k}/2)$ across qubits, so the state is a non-linear function $\phi(x)$ of the input features. Because a rotation is periodic in its angle, the kernel and variational models min-max scale each feature on the training sites before encoding, to $[0,\pi]$ and $[0,\pi/2]$ respectively, so that distinct feature values map to distinct states. To capture dependencies between features, we add controlled-NOT (CNOT) gates between qubits,
\begin{equation}
\mathrm{CNOT}\,|q_1 q_2\rangle = |q_1,\; q_1 \oplus q_2\rangle,
\end{equation}
where $q_1$ is the control qubit, $q_2$ the target qubit, and $\oplus$ denotes addition modulo two; the gate ties the qubits into a joint state that cannot be factored into independent single-qubit descriptions.

\subsection{Quantum Fidelity as a Connectivity Metric}

The design evaluated in this paper defines graph connectivity by the \emph{fidelity} between quantum-encoded regional representations, which are themselves derived from correlation-based features. The fidelity between two pure quantum states $|\psi_i\rangle$ and $|\psi_j\rangle$, representing two brain regions, is the squared overlap:

\begin{equation}
A_{ij}^{\text{quantum}} = |\langle \psi(x_i) | \psi(x_j) \rangle|^2
\label{eq:fidelity}
\end{equation}

Three properties of this metric matter for graph construction. Fidelity is equivalent to a quantum kernel $K(x_i, x_j)$, and because the feature map involves non-linear trigonometric terms, it defines a non-linear similarity function, whose relationship to linear and RBF kernels we assess empirically through matched performance, effective dimension, and eigenspectrum. Fidelity also measures the geometric similarity of the encoded states, a non-linear function of all of a region's features at once. Finally, in high-dimensional Hilbert spaces, most pairs of states are close to orthogonal, a consequence of measure concentration, so fidelity can give a sparser adjacency than a dense correlation matrix. Sparsity does not establish denoising, however: concentration pushes every overlap toward zero whether or not the pair is related, so it can suppress informative relationships along with noise. 

\subsection{Adaptive Graph Attention Mechanism}

The fidelity matrix $A^{\text{quantum}}$ supplies only a candidate topology: it is computed without reference to diagnostic labels, so it need not reflect the structure relevant to classification. We therefore use it to select candidate edges and let a graph attention network (GAT) \cite{velickovic2018gat} learn how strongly each contributes: in the principal experiments, the fidelity graph defines a fixed support that remains unchanged during training, while the edge weights and graph-classification parameters are optimised.

A standard GAT computes attention coefficients from node features alone,
\begin{equation}
\alpha_{ij} = \text{softmax}_j \big(\text{LeakyReLU}(a^{\top}[\Theta h_i \,\|\, \Theta h_j])\big),
\end{equation}
where $h_i$ is the feature vector of node $i$, $\Theta$ a shared learnable weight matrix, $a$ a learnable attention vector and $\|$ denotes concatenation. We modify this mechanism so that attention is restricted to the fidelity-selected candidate edges and combines three terms. Let $z_i$ be the latent vector of region $i$, given by the per-qubit Pauli-$Z$ expectation values of its encoded state; $\mathrm{Sim}(z_i,z_j)$ the similarity of the latents of regions $i$ and $j$; $\mathrm{Learnable}(z_i,z_j)$ a learned function of the pair; $\mathrm{Attn}_{\text{GAT}}(z_i,z_j)$ the standard attention score above; $\sigma(\cdot)$ a pointwise non-linearity; and $\alpha$, $\beta$ and $\gamma$ learnable mixing weights. The hybrid attention coefficient is

\begin{equation}
\begin{split}
\alpha_{ij}^{\text{hybrid}} = \text{softmax}_j \Big( \sigma \big( &\alpha \cdot \text{Sim}(z_i, z_j) \\
&+ \beta \cdot \text{Learnable}(z_i, z_j) \\
&+ \gamma \cdot \text{Attn}_{\text{GAT}}(z_i, z_j) \big) \Big)
\end{split}
\end{equation}

The mixing weights $\alpha$, $\beta$ and $\gamma$ are distinct from the double-indexed attention coefficients $\alpha_{ij}$ and from the RBF kernel width $\gamma$ used for the classical kernels. In training, $\mathrm{Sim}$ is the cosine similarity of the latents and $\mathrm{Learnable}$ is a small feed-forward network on the concatenated pair; their weighted sum, passed through a rectified linear unit (ReLU), forms the edge weight, which enters the GAT attention alongside the standard score.

\section{Methodology: The Architecture Under Test}\label{sec-methods}

\subsection{Dataset and Preprocessing}
\label{sec:dataset}

The architecture is developed on the Autism Brain Imaging Data Exchange, ABIDE~I \cite{DiMartino2014ABIDE,Craddock2013NBI}, which contributes $1{,}035$ subjects ($505$ ASD, $530$ controls) across $20$ acquisition sites after quality control. It is then evaluated on three further cohorts in Section~\ref{sec:multicohort}: ADHD-200 \cite{adhd200consortium2012} for attention-deficit/hyperactivity disorder (ADHD), REST-meta-MDD \cite{Yan2019RESTmetaMDD} for major depressive disorder (MDD), and UCLA-CNP \cite{Poldrack2016UCLACNP}, for $4{,}308$ subjects in total (Table~\ref{tab:cohorts}). The cohorts are reported separately and never pooled: their releases disagree on parcellation width ($200$ CC200 regions for ABIDE and REST-meta-MDD against $190$ for the others) and they were preprocessed with different pipelines (the Configurable Pipeline for the Analysis of Connectomes (C-PAC), Athena, the Data Processing \& Analysis for Brain Imaging (DPABI) toolbox, and our own), so pooling would confound preprocessing with site. The heterogeneity across scanner types, acquisition protocols, and demographics \emph{within} each cohort is what the leave-site-out protocol of Section~\ref{sec:design} is designed to stress-test. Fold counts reflect sites carrying both classes, and Table~\ref{tab:cohorts} reports the subjects contained in these folds. This affects ADHD-200 alone, whose release holds $767$ subjects, of which $619$ fall in sites carrying both classes. The other $148$ come from two control-only sites, which cannot serve as test folds because sensitivity, specificity, and AUC are undefined on a single class; they are used for training in every fold and never tested, so no test label is exposed and every quantum and classical model trains on the same subjects. The remaining three cohorts are used in full. UCLA-CNP has only two scanners, and a paired test on two folds cannot reach significance at any effect size, so it is evaluated with stratified 10-fold cross-validation; its folds are therefore within-site and its accuracies are not on the same scale as those of the other three. These four were selected because each releases per-subject diagnostic labels together with scanner or site identifiers, which the matched leave-site-out design requires; they are not a census of available rs-fMRI data.

\begin{table}[!tbp]
\caption{Cohorts evaluated, diagnostic groups, and brain networks reported as disrupted.}
\label{tab:cohorts}
\centering
\footnotesize
\setlength{\tabcolsep}{3pt}
\renewcommand{\arraystretch}{1.05}

\begin{tabularx}{\columnwidth}{|>{\raggedright\arraybackslash}p{1.35cm}|
>{\raggedright\arraybackslash}p{1.30cm}|
c|c|c|
>{\raggedright\arraybackslash}X|}
\hline

\textbf{Cohort} &
\textbf{Diagnostic groups} &
$\bm{N}$ &
\textbf{Folds} &
\textbf{ROIs} &
\textbf{Networks reported disrupted} \\
\hline

ABIDE~I &
ASD &
$1{,}035$ &
$20$ &
$200$ &
DMN $\downarrow$, frontoparietal $\downarrow$, limbic/amygdala $\downarrow$ \\
\hline

ADHD-200 &
ADHD &
$619$ &
$7$ &
$190$ &
frontoparietal $\downarrow$, DMN (delayed suppression), subcortical/striatal $\downarrow$ \\
\hline

REST-meta-MDD &
MDD &
$2{,}428$ &
$25$ &
$200$ &
DMN (direction contested), salience $\uparrow$, frontoparietal $\downarrow$ \\
\hline

UCLA-CNP &
SCZ, BD, ADHD, HC &
$226$ &
$10^{*}$ &
$190$ &
frontoparietal $\downarrow\downarrow$, thalamocortical, auditory $\uparrow$ \\
\hline

\end{tabularx}
\\[1pt]
{\scriptsize
$^{*}$Stratified 10-fold; all other cohorts are leave-site-out. For UCLA-CNP, all patients ($124$) are classified against HC ($102$). DMN: default-mode network; SCZ: schizophrenia; BD: bipolar disorder; HC: healthy controls.
}

\end{table}

\begin{figure*}[!t]
\centering
\includegraphics[width=0.44\textwidth]{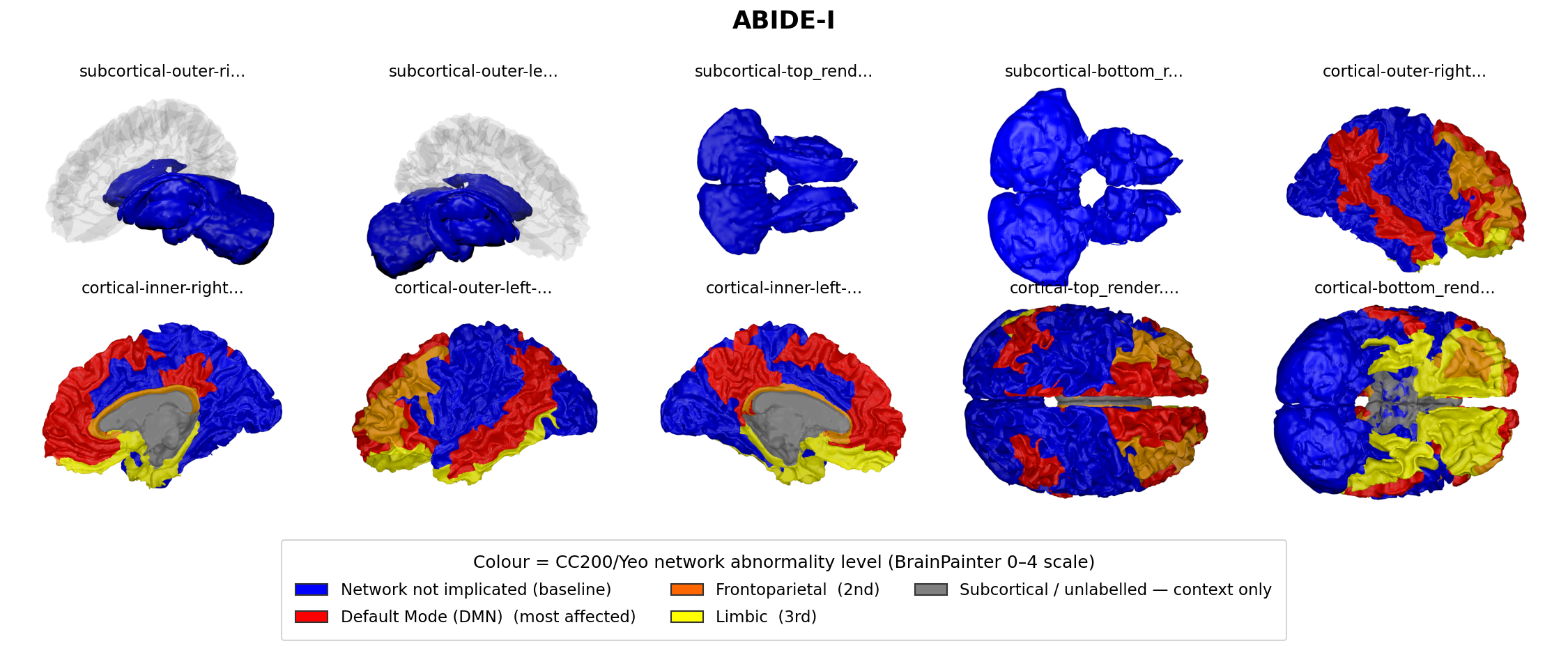}\hfill
\includegraphics[width=0.44\textwidth]{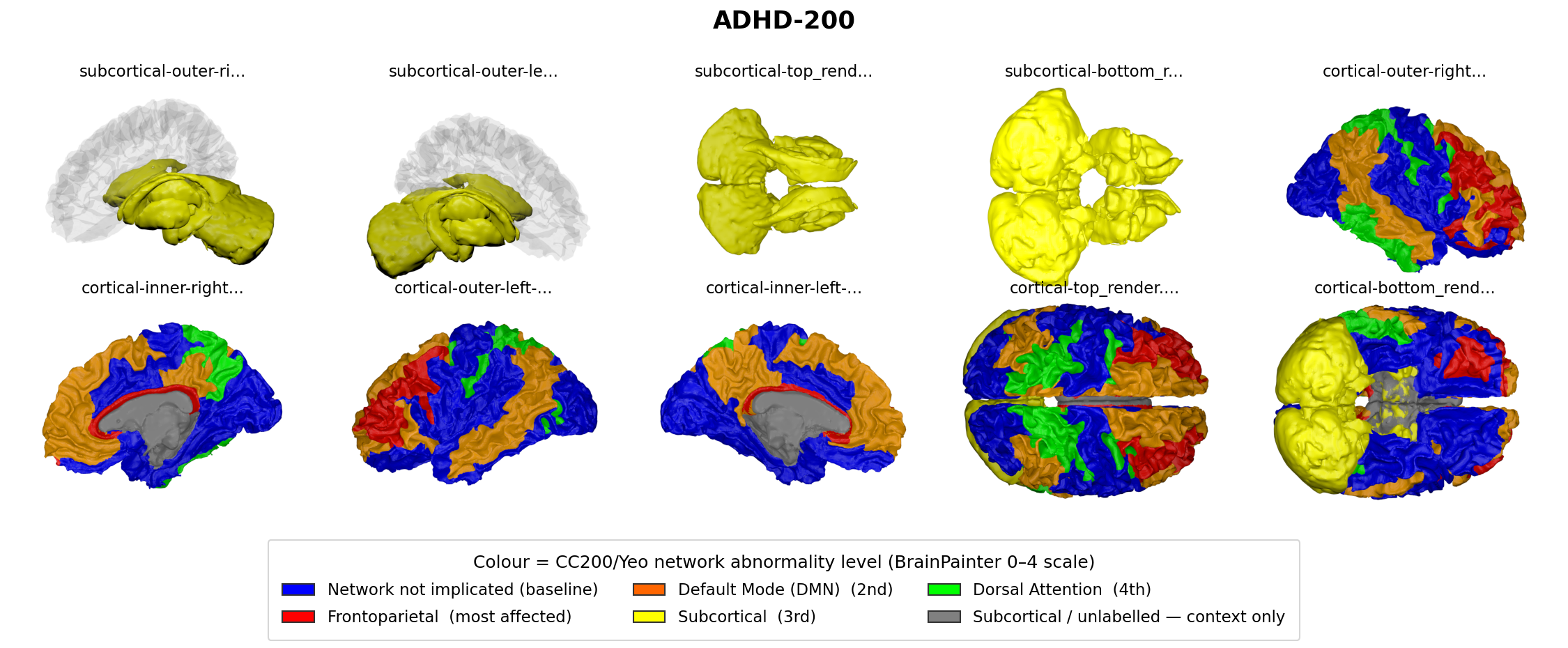}\\[4pt]
\includegraphics[width=0.44\textwidth]{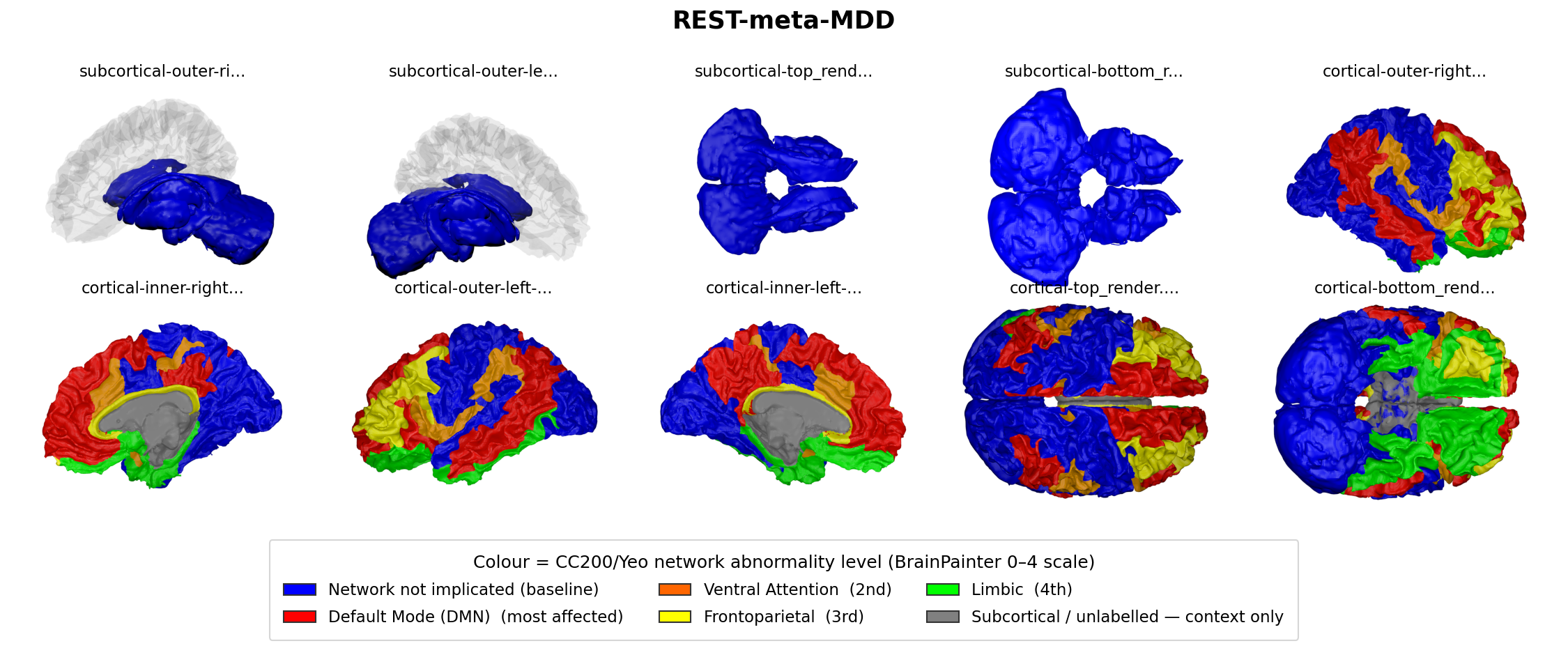}\hfill
\includegraphics[width=0.44\textwidth]{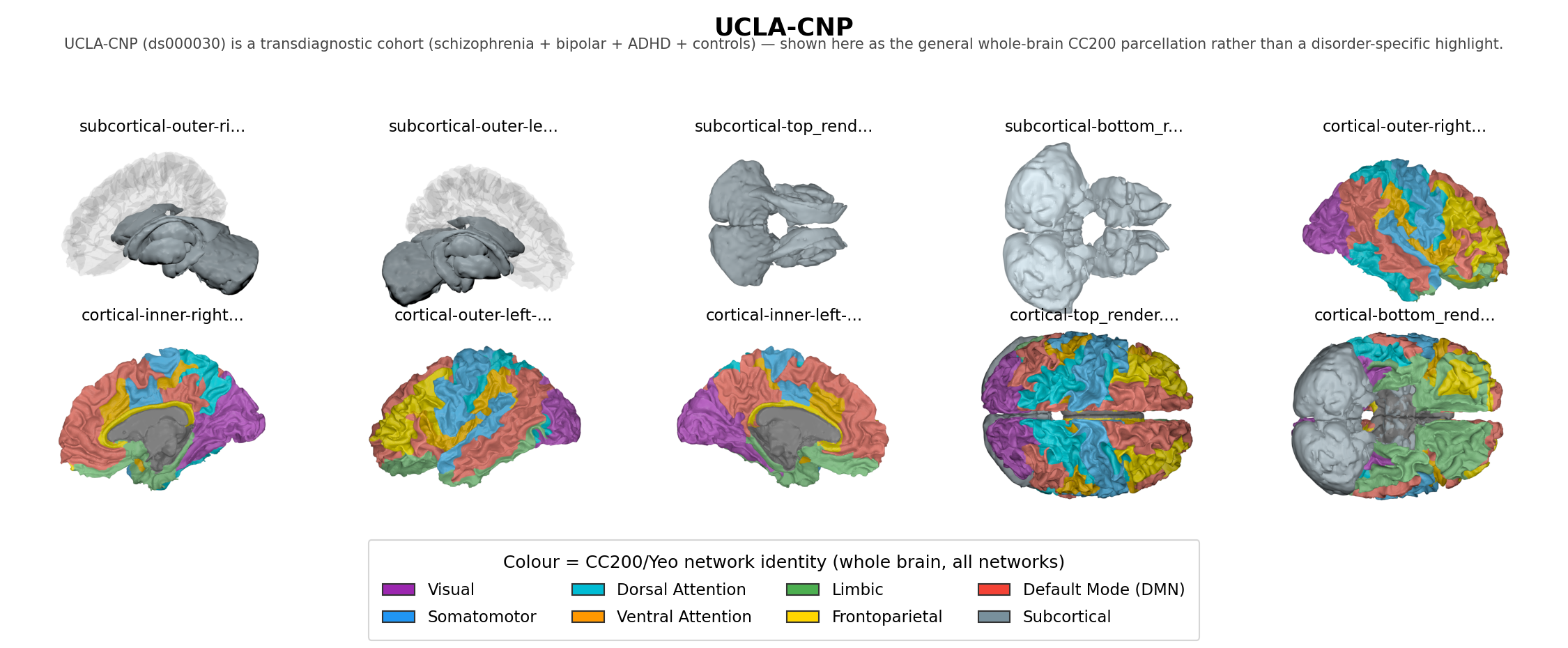}
\caption{Functional networks reported as disrupted in (a) ABIDE~I, (b) ADHD-200, (c) REST-meta-MDD and (d) UCLA-CNP.}
\label{fig:networks}
\end{figure*}

For ABIDE~I we use the version preprocessed with C-PAC: slice-timing and motion correction, nuisance regression over white matter, cerebrospinal fluid (CSF) and 24 motion parameters, bandpass filtering ($0.01$--$0.1$\,Hz), and parcellation into 200 CC200 regions covering the cerebrum and cerebellum. All four cohorts are represented on the CC200 functional parcellation \cite{craddock2012}, whose regions are data-driven functional clusters rather than named anatomical structures. The atlas specifies $200$ clusters, but the spatially constrained spectral clustering resolves to $190$ non-empty parcels, which is why releases disagree on width.

\noindent\textbf{Output:} for each subject, a feature matrix $X_{\text{raw}} \in \mathbb{R}^{200 \times T}$, where $T$ ranges from 150 to 300 depending on site.

\noindent\textbf{Dimensionality reduction:} current noisy intermediate-scale quantum (NISQ) devices and simulators have limited qubit capacity for efficient gradient computation, and directly encoding $T \approx 200$ timepoints would need prohibitively deep circuits. We therefore reduce each region's description to $n_\text{qubits}$ components, eight in the reported configuration. In every variant, the matrix presented to principal component analysis (PCA) has the $200$ regions as its observations, so the fit is per subject or per fold and never per region.

Two descriptions of a region and two fitting scopes are used. The first fits PCA on $X_{\text{raw}} \in \mathbb{R}^{200 \times T}$, one fit per subject, with each region's time course standardised beforehand so that components follow temporal shape rather than BOLD amplitude; it retains $59.9\%$ of the variance. The second describes region $i$ by its connectivity profile, row $i$ of the subject's $200 \times 200$ correlation matrix, and fits one basis on the stacked profiles of every training-site subject, a $(200 N_{\text{train}}) \times 200$ matrix with $N_{\text{train}}$ the number of training subjects, applying it unchanged to the held-out site; it retains $91.1\%$. Fitting inside the fold preserves the site-held-out guarantee. Measured with no graph involved, these two choices, what a node describes and where its basis is fitted, move held-out AUC from $0.459$ to $0.739$, further than any quantum--classical difference measured in this study. The quantum model suite of Section~\ref{sec:quantum-suite} does not use this compression; it consumes the supervised connectivity features described below.

\noindent\textbf{Final input:} a node feature matrix $X \in \mathbb{R}^{200 \times d}$ with $d = n_\text{qubits}$, so each brain region is described by $d$ latent features, with $d=8$ in the reported configuration and $d=16$ in the width study of Section~\ref{sec:width}.

\noindent\textbf{Supervised feature selection:}\label{sec:supervised-features} Unsupervised PCA maximises variance rather than class separation, so it can discard precisely the directions that distinguish groups, and published pipelines on these cohorts generally use supervised selection instead. Every matched quantum--classical comparison therefore uses the top-$k$ connections ranked by absolute $t$-statistic ($k=8$ for the matched models, $k=2{,}000$ for the unrestricted ones), fitted on the training sites only. Each matched model thus receives Pearson correlation coefficients rather than BOLD time series, so quantum and classical models see identical inputs and differ only in the similarity they apply. Selection is performed inside each fold because ranking features on the full cohort would leak held-out labels into the feature stage, a form of leakage shown to inflate connectome-based prediction performance \cite{rosenblatt2024leakage}.

\subsection{System Architecture Overview}
\label{sec:architecture}

\begin{figure*}[!t]
\centering
\includegraphics[width=0.90\textwidth]{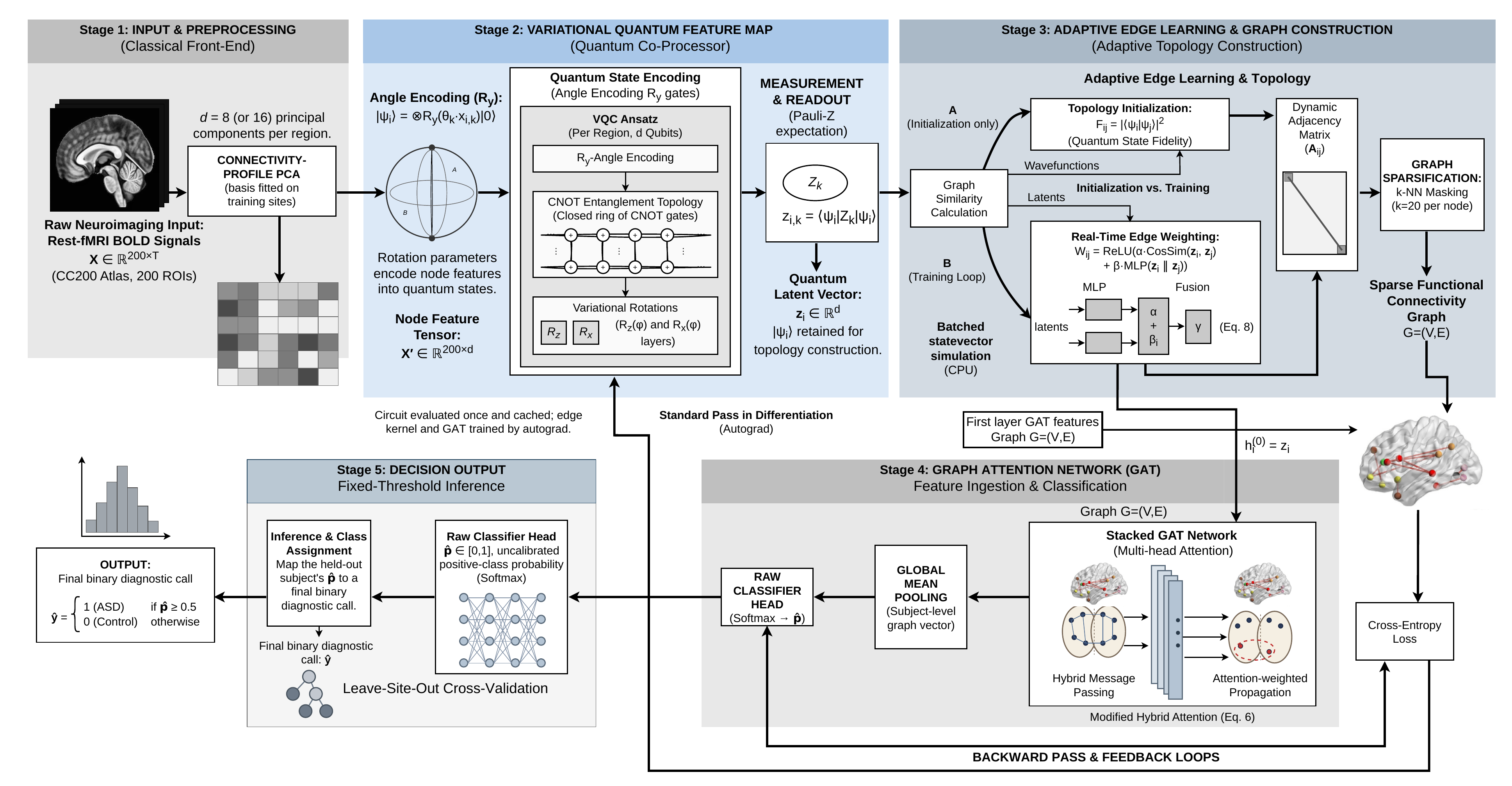}
\caption{The fidelity-graph architecture evaluated here, showing quantum feature encoding and adaptive graph construction.}
\label{fig:pipeline}
\end{figure*}

The evaluated system is a hybrid pipeline: a simulated quantum circuit handles state preparation and fidelity computation, and a classical processor handles edge learning and graph classification, as shown in Fig.~\ref{fig:pipeline}. A quantum feature map generates node representations and graph support, which are subsequently processed by a classical GAT.

\textbf{The Quantum Feature Encoder} maps regional connectivity features into quantum states in a Hilbert space. Circuits are simulated with a batched PyTorch statevector simulator, with PennyLane as the reference implementation (Section~\ref{sec:training}). The register carries $n$ qubits per region, one per retained PCA component; the value of $n$ is determined through fidelity-density and computational-cost analyses. Fidelity at $n=16$ leaves too few usable pairs at three of the four floors tested (Section~\ref{sec:fidelity-collapse}), while wider variational circuits are affordable only up to a point (Section~\ref{sec:width}). Every comparison in this paper uses $n=8$, which provides sufficient connectivity density while remaining computationally practical.

\begin{figure}[!tbp]
\centering
\includegraphics[width=\columnwidth]{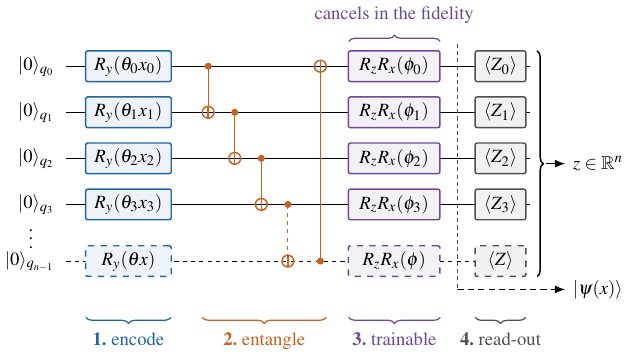}
\caption{The per-region circuit, shown for four of $n$ qubits.}
\label{fig:circuit}
\end{figure}

Fig.~\ref{fig:circuit} shows the per-region circuit gate by gate: a layer of $R_y$ rotations whose angles carry the input, a ring of CNOTs closing from $q_{n-1}$ back to $q_0$, a trainable $R_z R_x$ layer, and a per-qubit Pauli-$Z$ read-out. The encoding layer prepares
\begin{equation}
|\psi(x_i)\rangle = \bigotimes_{k=1}^{n} R_y(\theta_k \cdot x_{i,k}) |0\rangle
\label{eq:encoding}
\end{equation}
where $\theta_k$ is the trainable gain on the $k$-th encoding rotation. In the principal experiments, all gains remained fixed at their initial value of one. After the CNOT ring and the variational $R_z R_x$ layer (angles $\phi_k$), a per-qubit Pauli-$Z$ measurement returns a latent vector $z_i \in \mathbb{R}^{n}$ with entries $z_{i,k} = \langle\psi_i|Z_k|\psi_i\rangle$, where $|\psi_i\rangle$ is the final circuit state; this vector forms the node features, with the full state vector retained separately for the fidelity of Eq.~\eqref{eq:fidelity}. In that fidelity, every data-independent unitary applied after the encoding, the CNOT ring as well as the variational layer, cancels from the overlap, because it acts identically on both states (Section~\ref{sec:invariance}). The encoder carries no decoder and no reconstruction objective; in the principal graph runs, it is kept at its initial parameters (Section~\ref{sec:training}).

\textbf{The Adaptive Edge Learning Module} learns edge weights over a graph support that fidelity establishes once at initialisation, recomputing the weights at every training step.

\begin{itemize}
\item \textbf{Input:} the collection of latent quantum vectors $\{z_1, \dots, z_{200}\}$.
\item \textbf{Mechanism:} quantum fidelity (Eq.~\eqref{eq:fidelity}) selects the candidate edges: the squared overlap of two regions' states ranks how strongly they are connected, and the strongest pairs form the support.

During training, the edge weights on this support are computed from the latent representations rather than recalculated from state fidelity:
\begin{equation}
W_{ij} = \text{ReLU} \left( \alpha \cdot \text{CosSim}(z_i, z_j) + \beta \cdot \text{MLP}(z_i \| z_j) \right)
\label{eq:edge}
\end{equation}
Here $\alpha$ and $\beta$ are learnable scalars, $\mathrm{CosSim}(u,v)=u^{\top}v/(\lVert u\rVert\,\lVert v\rVert)$ is cosine similarity, $\mathrm{MLP}$ is a learned multilayer perceptron (MLP) and $\|$ denotes concatenation.
\item \textbf{Sparsification:} to control noise and keep the GAT's computation manageable, we apply $k$-nearest-neighbour masking, keeping only the top $k=20$ connections per node, 10\% of the 200 regions, which keeps the graph sparse while leaving each node a choice of neighbours for attention to weight. The result is left \emph{directed} and is not symmetrised: region $i$ may retain an edge to $j$ without the converse, since $j$ may have twenty stronger neighbours of its own. Attention is computed over incoming edges, so the asymmetry is well defined.
\end{itemize}

\textbf{The Graph Attention Network (GAT)} performs the binary classification.
Each node's quantum latent becomes its initial GAT feature, $h_i^{(0)} = z_i$, so gradients from the classification loss reach the classical attention layers and the adaptive-edge kernel. The resulting adjacency, $A^{\text{quantum}}$ in its support and $W$ of Eq.~\eqref{eq:edge} in its values, is passed to the GAT for propagation:

\begin{equation}
h_i' = \alpha_{i,i}\Theta h_i + \sum_{j \in \mathcal{N}(i)} \alpha_{i,j}\Theta h_j
\end{equation}

where $h_i'$ is the updated feature of node $i$, $\mathcal{N}(i)$ its set of incoming neighbours in the sparsified graph and $\alpha_{i,j}$ the attention coefficients, modulated by the edge weights of Eq.~\eqref{eq:edge} on the fidelity-selected support.

The classifier is a two-layer GAT. The first layer maps each region's $n=8$-dimensional latent to $32$ features in each of four attention heads, concatenated to $128$; the second maps these to $32$ features with a single head. Both use exponential linear unit (ELU) activations and dropout $p=0.6$, chosen to limit overfitting given the sample size. Global mean pooling gives a graph-level embedding, which a linear layer maps to the two classes with a softmax read-out.

\subsection{Training Setup}
\label{sec:training}
Only the classical components are trained. Before training, the PCA basis is fitted on the training sites, the circuit is evaluated once for every region, and the latents and the fidelity-selected support are cached; each training step then passes the cached latents through edge learning and the GAT to the cross-entropy loss $\mathcal{L}=-\sum_{c=1}^{2} y_c\log\hat{y}_c$, where $y_c$ is the one-hot label and $\hat{y}_c$ the predicted class probability. Gradients from PyTorch autograd update the GAT, the edge-learning MLP, and $\alpha,\beta$ with Adam \cite{kingma2015adam} (learning rate $0.01$, weight decay $5\times10^{-4}$, $30$ epochs). In the principal graph runs (all rows of Table~\ref{tab:results-graph} but the trained-encoder variant), the quantum encoder, with every encoding gain equal to one, and the fidelity-derived topology therefore remain fixed throughout training, while only the classical edge-learning and GAT parameters are updated; the circuits of the model suite (Section~\ref{sec:quantum-suite}) are trained separately. For the kernel models, bandwidth is selected by kernel-target alignment (KTA) \cite{cristianini2002kta} on the training sites of each fold, with no inner cross-validation loop, and reported metrics are averaged across folds. Algorithm~\ref{alg:pipeline} summarises one training step.

\noindent\textbf{Implementation:} Experiments used Python 3.13, with PyTorch \cite{paszke2019pytorch} for a batched statevector simulator, PyTorch Geometric for graph operations, and PennyLane as the reference quantum backend. Because the topology touches all $19{,}900$ region pairs per subject, the simulator evolves every region of a subject as one batched tensor, and circuits that are trained take gradients through autograd; PennyLane, which executes region by region with adjoint differentiation, agrees with it to simulator precision but is too slow for the fold counts reported here. All experiments (four cohorts, six architectural families, $62$ folds) ran on one laptop (Apple M4 Pro, 14 cores, 24\,GB memory, CPU only); no circuit exceeded sixteen qubits, and the largest state vector held in memory was $0.16$\,GB, so no specialised hardware is needed to reproduce the results.

\begin{algorithm}[!tbp]
\caption{One hybrid training step on a fixed support}
\label{alg:pipeline}
\small
\begin{algorithmic}[1]
\REQUIRE BOLD series $X_{\text{raw}} \in \mathbb{R}^{200\times T}$ and label $y$ of one subject; basis $P$ fitted on the training sites of the fold; encoding gains $\{\theta_k\}$, variational angles $\{\phi_k\}$, edge weights $(\alpha,\beta)$, GAT weights $\Theta$
\ENSURE parameters after one gradient step
\STATE $X \leftarrow P\big(\mathrm{profiles}(X_{\text{raw}})\big)$ \COMMENT{$n=8$ components}
\FOR[once, then cached]{each region $i = 1, \dots, 200$}
\STATE prepare $|\psi(x_i)\rangle$: $R_y(\theta_k x_{i,k})$, CNOT ring, $R_z R_x$ layer
\STATE $z_i \leftarrow \big(\langle\psi(x_i)|Z_k|\psi(x_i)\rangle\big)_{k=1}^{n}$
\ENDFOR
\IF[once, without gradient]{the topology is not yet initialised}
\STATE $F_{ij} \leftarrow |\langle\psi(x_i)|\psi(x_j)\rangle|^2$ for all pairs \COMMENT{Eq.~\eqref{eq:fidelity}}
\STATE keep the $k=20$ highest-fidelity neighbours of each region \COMMENT{discrete top-$k$}
\ENDIF
\FOR{each retained edge $(i,j)$}
\STATE $W_{ij} \leftarrow \mathrm{ReLU}\big(\alpha\,\mathrm{CosSim}(z_i,z_j) + \beta\,\mathrm{MLP}(z_i \Vert z_j)\big)$
\ENDFOR
\STATE $\hat{y} \leftarrow \mathrm{GAT}_\Theta(\{z_i\}, W)$ with global mean pooling
\STATE $\mathcal{L} \leftarrow \mathrm{CrossEntropy}(\hat{y}, y)$
\STATE gradients for $\Theta,\alpha,\beta$ by autograd \COMMENT{$\theta,\phi$ held at initial values}
\STATE update $\Theta,\alpha,\beta$ with Adam
\end{algorithmic}
\end{algorithm}

\subsection{Experimental Design and Evaluation Metrics}
\label{sec:design}

The design principle throughout is that a quantum model is compared against the classical model it corresponds to (same features, same folds, same optimiser, same training budget) rather than against the best classical number available anywhere; Section~\ref{sec:artefacts} describes the artefacts this matching prevents.

The evaluation protocol follows three rules.

\textbf{Leave-site-out (LSO) validation:} the dataset is split by acquisition site. In each fold, one site is held out entirely for testing while the rest train the model, which reduces the ability to exploit site-specific scanner characteristics, a well-known confound in ABIDE-based work. UCLA-CNP, with two scanners, is the exception (Section~\ref{sec:dataset}).

\textbf{Baselines:} linear classical (support vector machine (SVM) with a linear kernel; graph convolutional network (GCN) \cite{kipf2017gcn} with a Pearson-correlation graph) and non-linear classical (SVM with an RBF kernel; GCN with an RBF-kernel similarity graph), with both SVMs applied to flattened correlation matrices (Tables~\ref{tab:results-graph}--\ref{tab:results-reference}).

\textbf{Statistical testing:} every quantum and classical pair is compared with a paired Wilcoxon signed-rank test over identical folds, with $95\%$ confidence intervals (CIs) across folds reported for the ABIDE~I results of Tables~\ref{tab:results-graph}--\ref{tab:results-reference} and Bonferroni correction applied wherever a sweep of comparisons is reported. A label-permutation null is run for the kernel arm on ABIDE~I, shuffling labels within site so that per-site class prevalence is preserved and feature selection, scaling, and bandwidth choice are repeated inside every fold on the permuted labels (Section~\ref{sec:matched-comparison}).

Alongside accuracy, F1, AUC, and specificity, we report the Matthews correlation coefficient (MCC)
\begin{equation}
\mathrm{MCC} = \frac{TP \cdot TN - FP \cdot FN}{\sqrt{\hat{P}\,P\,\hat{N}\,N}},
\label{eq:mcc}
\end{equation}
where $TP$, $TN$, $FP$ and $FN$ count true positives, true negatives, false positives and false negatives, with $P = TP{+}FN$, $N = TN{+}FP$, $\hat{P} = TP{+}FP$ and $\hat{N} = TN{+}FN$,
which, unlike accuracy, F1 or specificity, stays near zero for a predictor that
assigns every subject to one class. Algorithm~\ref{alg:protocol} sets out the
matched protocol that every quantum--classical comparison follows.

\begin{algorithm}[!tbp]
\caption{Matched comparison of a quantum model and its classical twin}
\label{alg:protocol}
\small
\begin{algorithmic}[1]
\REQUIRE cohort $\mathcal{D}$ with sites $1,\dots,S$; quantum model $Q$ and classical model $C$ matched on features, capacity, optimiser and training budget
\ENSURE per-fold metrics, paired $p$-values, screened wins
\FOR{each site $s$ carrying both classes \COMMENT{UCLA-CNP: stratified 10-fold}}
\STATE $\mathcal{D}_{\text{tr}} \leftarrow \mathcal{D}\setminus s$; \ $\mathcal{D}_{\text{te}} \leftarrow s$
\STATE $\mathcal{S} \leftarrow \mathrm{select}(\mathcal{D}_{\text{tr}}, n)$, \ $P \leftarrow \mathrm{basis}(\mathcal{D}_{\text{tr}})$; apply $P \circ \mathcal{S}$ unchanged to $\mathcal{D}_{\text{te}}$ \COMMENT{no fit sees $\mathcal{D}_{\text{te}}$}
\STATE kernel models: $\gamma^{\ast} \leftarrow \arg\max_{\gamma} \mathrm{KTA}\big(K_{\gamma}(\mathcal{D}_{\text{tr}}), yy^{\top}\big)$ \COMMENT{Eq.~\eqref{eq:kta}, on $\mathcal{D}_{\text{tr}}$}
\STATE train $Q$ and $C$ on $\mathcal{D}_{\text{tr}}$ with the same seed and budget
\STATE record accuracy, F1, AUC, sensitivity, specificity and MCC of both on $\mathcal{D}_{\text{te}}$
\ENDFOR
\FOR{each metric $m$}
\STATE $p_m \leftarrow$ paired Wilcoxon signed-rank test on the per-fold differences $m(Q)-m(C)$
\ENDFOR
\STATE mark $Q$ or $C$ degenerate if, pooled over folds, its sensitivity or specificity is below $0.30$ or $|\mathrm{MCC}|<0.02$
\STATE report each $p_m$ uncorrected and against $0.05/M$ for the $M$ tests of the sweep; count a win as clean only if $p_m<0.05$ and neither $Q$ nor $C$ is degenerate
\end{algorithmic}
\end{algorithm}

\section{Experimental Results}\label{sec-results}

\subsection{Classification Performance}

Tables~\ref{tab:results-graph}--\ref{tab:results-reference} report performance averaged across the $20$ ABIDE~I site folds. Holding out an entire scanner lowers absolute accuracy relative to a random split but estimates generalisation to an unseen site. The three tables answer different questions and are not interchangeable: \emph{graph-based} models build a topology over $200$ regions
and classify the graph, \emph{feature-based} models (Table~\ref{tab:results-matched}; defined in Section~\ref{sec:quantum-suite}) receive the eight supervised-selected connectivity features that the register admits (Section~\ref{sec:supervised-features}), and the \emph{reference} models take the full correlation vector with no register limit. F1 and specificity treat the disorder group as positive. MCC is the column that separates classification from degeneracy: it stays at zero for a predictor that assigns every subject to one class, where F1 reaches $\approx0.65$ on a near-balanced cohort.

\begin{table}[!tbp]
\caption{Leave-site-out performance of the graph models on ABIDE~I ($1{,}035$ subjects, $20$ site folds, eight qubits; mean $\pm$ 95\% CI, normal quantile).}
\label{tab:results-graph}
\centering
\scriptsize
\setlength{\tabcolsep}{1.0pt}
\renewcommand{\arraystretch}{1.05}
\resizebox{\columnwidth}{!}{%
\begin{tabular}{|l|c|c|c|c|}
\hline
\textbf{Model} & \textbf{F1} & \textbf{Acc} & \textbf{Spec} & \textbf{MCC} \\
\hline
GCN+Pearson & $0.252${\tiny$\pm0.118$} & $0.494${\tiny$\pm0.029$} & $0.664${\tiny$\pm0.185$} & $-0.007${\tiny$\pm0.036$} \\
GCN+RBF & $0.145${\tiny$\pm0.119$} & $0.508${\tiny$\pm0.029$} & $0.783${\tiny$\pm0.179$} & $-0.009${\tiny$\pm0.017$} \\
Quantum+GAT & $0.196${\tiny$\pm0.135$} & $0.504${\tiny$\pm0.028$} & $0.700${\tiny$\pm0.206$} & $0.000${\tiny$\pm0.000$} \\
Quantum+GAT, trained enc. & $0.237${\tiny$\pm0.130$} & $0.513${\tiny$\pm0.036$} & $0.705${\tiny$\pm0.174$} & $0.010${\tiny$\pm0.055$} \\
Quantum+GCN (abl.) & $0.133${\tiny$\pm0.119$} & $0.512${\tiny$\pm0.027$} & $0.800${\tiny$\pm0.180$} & $0.000${\tiny$\pm0.000$} \\
\hline
\end{tabular}}
\\[1pt]
{\scriptsize For contrast, the unrestricted linear SVM of
Table~\ref{tab:results-reference} reaches F1 $0.653$, accuracy $0.671$ and MCC
$0.346$ on the same folds.}
\end{table}

\begin{table}[!tbp]
\caption{Matched eight-feature comparisons on ABIDE~I, each quantum model ($\dagger$) above its classical twin (mean $\pm$ 95\% CI over $20$ site folds).}
\label{tab:results-matched}
\centering
\scriptsize
\setlength{\tabcolsep}{1.0pt}
\renewcommand{\arraystretch}{1.05}
\resizebox{\columnwidth}{!}{%
\begin{tabular}{|l|c|c|c|c|}
\hline
\textbf{Model} & \textbf{Acc} & \textbf{AUC} & \textbf{Spec} & \textbf{MCC} \\
\hline
QCNN$\dagger$ & $0.615${\tiny$\pm0.029$} & $0.666${\tiny$\pm0.036$} & $0.699${\tiny$\pm0.076$} & $0.253${\tiny$\pm0.056$} \\
CNN (1-D) & $0.589${\tiny$\pm0.038$} & $0.638${\tiny$\pm0.043$} & $0.652${\tiny$\pm0.074$} & $0.194${\tiny$\pm0.075$} \\
\hline
VQC$\dagger$ & $0.582${\tiny$\pm0.032$} & $0.612${\tiny$\pm0.043$} & $0.632${\tiny$\pm0.074$} & $0.179${\tiny$\pm0.066$} \\
MLP & $0.600${\tiny$\pm0.028$} & $0.629${\tiny$\pm0.037$} & $0.674${\tiny$\pm0.055$} & $0.208${\tiny$\pm0.057$} \\
\hline
QSVM-fixed$\dagger$ & $\mathbf{0.622}${\tiny$\pm\mathbf{0.034}$} & $0.663${\tiny$\pm0.041$} & $0.717${\tiny$\pm0.076$} & $\mathbf{0.264}${\tiny$\pm\mathbf{0.070}$} \\
SVM (RBF), matched & $0.616${\tiny$\pm0.030$} & $0.662${\tiny$\pm0.039$} & $0.702${\tiny$\pm0.075$} & $0.255${\tiny$\pm0.059$} \\
\hline
TQEK$\dagger$ & $0.610${\tiny$\pm0.036$} & $\mathbf{0.686}${\tiny$\pm\mathbf{0.036}$} & $\mathbf{0.724}${\tiny$\pm\mathbf{0.069}$} & $0.234${\tiny$\pm0.077$} \\
Trainable-RBF & $0.581${\tiny$\pm0.036$} & $0.606${\tiny$\pm0.044$} & $0.586${\tiny$\pm0.062$} & $0.177${\tiny$\pm0.074$} \\
\hline
\end{tabular}}
\end{table}

\begin{table}[!tbp]
\caption{Unrestricted classical references on ABIDE~I: full correlation vector, no register limit.}
\label{tab:results-reference}
\centering
\scriptsize
\setlength{\tabcolsep}{1.0pt}
\renewcommand{\arraystretch}{1.05}
\resizebox{\columnwidth}{!}{%
\begin{tabular}{|l|c|c|c|c|}
\hline
\textbf{Model} & \textbf{F1} & \textbf{Acc} & \textbf{Spec} & \textbf{MCC} \\
\hline
SVM (Linear), full corr. & $\mathbf{0.653}${\tiny$\pm\mathbf{0.039}$} & $\mathbf{0.671}${\tiny$\pm\mathbf{0.036}$} & $0.689${\tiny$\pm0.067$} & $\mathbf{0.346}${\tiny$\pm\mathbf{0.073}$} \\
SVM (RBF), full corr. & $0.627${\tiny$\pm0.054$} & $0.642${\tiny$\pm0.042$} & $0.634${\tiny$\pm0.088$} & $0.305${\tiny$\pm0.084$} \\
\hline
\end{tabular}}
\end{table}

Three observations emerge. First, the largest performance difference was observed between graph-based and correlation-feature classifiers rather than between quantum and classical methods: both SVMs separate the classes ($0.671$ and $0.642$ accuracy), while all five graph configurations sit at chance ($0.494$ to $0.513$), and the quantum convolutional network reaches $0.615$ against $0.589$ for its matched classical counterpart. Second, additional kernel non-linearity did not improve performance under the evaluated feature representations and validation protocol: the linear SVM beats the RBF SVM ($0.653$ against $0.627$ F1), and GCN on Pearson beats GCN on RBF ($0.252$ against $0.145$). Third, quantum topology did not outperform correlation topology under the evaluated datasets and experimental settings: Quantum+GAT reaches F1 $0.196$ against $0.252$ for GCN+Pearson, with accuracies within $0.010$ of each other and of chance.
Specificity is the only column where the quantum configurations appear competitive, but the highest specificities belong to the worst classifiers: the quantum-GCN ablation attains $0.800$ with an F1 of only $0.133$. Per-fold inspection shows folds collapsing onto a single predicted class, which the MCC column states directly: both fidelity-graph configurations sit at $0.000 \pm 0.000$, and only the correlation-based references clearly exceed zero.

Supervised selection is consequential in two ways. Relative to PCA, held-out AUC on ABIDE~I rises from $0.729$ to $0.738$, and, more importantly, the quantum comparison changes character: what appeared to be a significant quantum deficit under PCA reaches statistical parity with the classical result once the features are chosen for discriminability.

\subsection{Matched-Feature Comparison}
\label{sec:matched-comparison}

A quantum kernel restricted to $n$ qubits sees $n$ features, while an unrestricted classical model sees $2{,}000$. Comparing them directly measures dimensionality rather than the kernel. We therefore add a \emph{matched} control: a classical RBF kernel, $k_\gamma(x,x')=\exp(-\gamma\lVert x-x'\rVert^{2})$ with width parameter $\gamma$, given exactly the $n_\text{qubits}$ features the
quantum kernel receives.

\begin{figure}[!tbp]
\centering
\includegraphics[width=\columnwidth]{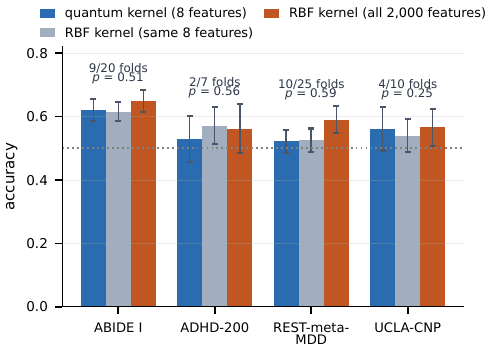}
\caption{Quantum kernel against classical kernels given matched and full feature sets on four cohorts.}
\label{fig:benchmark}
\end{figure}

Under the matched control, the quantum kernel and the classical kernel are statistically indistinguishable on every cohort: ABIDE~I $9/20$ folds ($p=0.51$), ADHD-200 $2/7$ ($p=0.56$), REST-meta-MDD $10/25$ ($p=0.59$), UCLA-CNP $4/10$ ($p=0.25$). Across $62$ folds, no statistically significant differences were observed. Section~\ref{sec:bounds} bounds how large any difference can be. Fig.~\ref{fig:benchmark} shows the quantum kernel tracking a classical kernel given the same eight features, while both trail a classical model given all $2{,}000$, with each cohort's fold wins and paired $p$-value marked above its bars.

These findings suggest that the quantum-edge deficit in the population-graph comparison of Section~\ref{sec:multicohort} was an artefact of comparing eight quantum features against $128$-dimensional classical descriptors. Where the quantum kernel does lose to the full $2{,}000$-feature models on ABIDE and REST-meta-MDD, the cause is the register limit rather than the kernel.

\subsection{Permutation, Seed and Noise Robustness Checks}
\label{sec:robustness}

Three checks establish that the parity is between working models rather than between two failures, and that it does not depend on noiseless simulation. First, a label-permutation null: labels were shuffled within site, preserving each site's class prevalence, and the leave-site-out sweep was re-run $104$ times with feature selection, scaling, and bandwidth choice repeated inside every fold. Every observed accuracy and AUC exceeded all $104$ permutations, so the empirical $p$ reaches its floor of $0.0095$, below the Bonferroni threshold
of $0.0125$ for four models, and the nulls sit at chance to three decimals. 
QSVM-fixed reaches $0.622$ accuracy and $0.663$ AUC against nulls of $0.498 \pm 0.019$ and $0.499 \pm 0.022$; TQEK $0.610$ and $0.686$ against $0.499 \pm 0.024$ and $0.498 \pm 0.032$; the matched SVM (RBF) $0.616$ and $0.662$ against $0.499 \pm 0.025$ and $0.498 \pm 0.030$; and Trainable-RBF $0.581$ and $0.606$ against $0.500 \pm 0.017$ and $0.496 \pm 0.022$.

Second, the suite was repeated at four seeds. The largest standard deviation of mean accuracy is $0.017$, and the two kernel models without a stochastic component are identical at every seed. Three of the four matched differences hold their sign throughout (QCNN $+0.025$, TQEK $+0.026$, QSVM-fixed $+0.006$ over their twins); only VQC against the matched MLP changes sign. The consistency is in direction, not magnitude: none survives the correction of Section~\ref{sec:quantum-wins}, and the largest is smaller than the cost of the eight-feature budget itself.

Third, shot and device noise: with every kernel entry estimated from a finite number of measurement shots on the same ABIDE~I folds, QSVM-fixed reaches $0.593$ accuracy at $128$ shots and $0.622$--$0.643$ (AUC $0.654$--$0.667$) from $256$ to $8{,}192$ shots, against $0.622$ under exact simulation; a global depolarising channel of rate $0.05$--$0.4$ leaves it at $0.621$--$0.634$ (AUC $0.652$--$0.669$). No noisy setting falls significantly below the matched RBF kernel.

\subsection{Why the Two Kernels Perform Alike}
\label{sec:kernel-spectrum}

A Gram matrix's participation ratio, $(\sum_i \lambda_i)^2 / \sum_i \lambda_i^2$, with $\lambda_i$ the eigenvalues of the matrix, counts how many directions a kernel actually uses, and we refer to it as the effective dimension. Near $1$, the kernel is almost constant and every subject looks alike; near the number of subjects, it approaches the identity and every subject becomes its own cluster. Useful kernels lie in between.

\begin{figure}[!tbp]
\centering
\includegraphics[width=\columnwidth]{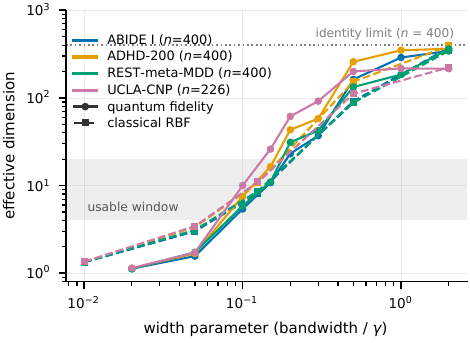}
\caption{Effective dimension of the quantum fidelity and classical RBF kernels against their width parameter on all four cohorts.}
\label{fig:kernel-spectrum}
\end{figure}

Measured on $400$ ABIDE subjects with eight selected features, both kernels sweep the same range as their width parameter varies, from an effective dimension of $1.1$ to $343$ out of a possible $400$
(Fig.~\ref{fig:kernel-spectrum}, blue). Both possess a usable window: quantum fidelity at bandwidth $0.15$ reaches effective dimension $10.8$, against $8.2$ for the RBF kernel at $\gamma=0.125$. At that matched setting, their eigenvalue spectra are nearly indistinguishable: the logarithms of the leading $40$ normalised eigenvalues correlate at $r = 0.997$, and those eigenvalues carry $87.0\%$ and $87.9\%$ of the total spectral weight.
The same overlap through the usable window holds on all four cohorts, as the remaining curves of Fig.~\ref{fig:kernel-spectrum} show.

This provides a mechanistic account of the matched performance in Fig.~\ref{fig:benchmark}. The quantum
feature map defines a non-linear similarity function that, on these data, behaves like a bandwidth-controlled RBF kernel: the two kernels span the same range of effective dimension, their leading eigenspectra nearly coincide once matched, and the encoding bandwidth of the quantum kernel plays the role that $\gamma$ plays for the RBF kernel. This empirical similarity does not prove that the two kernels belong to the same mathematical family, or that one is a reparameterisation of the other; it does, however, offer a plausible explanation for their comparable predictive performance.

It also clarifies why bandwidth was the decisive hyperparameter (Section~\ref{sec:matched-comparison}). An untuned bandwidth of $1.0$ places the quantum kernel at effective dimension $289$, close to the identity, where it memorises the training set and generalises poorly, which is where an
unreported bandwidth would leave it by default.

\section{Quantum Model Suite}
\label{sec:quantum-suite}

The matched comparisons of Section~\ref{sec:matched-comparison} turn on a single quantum construct, a fidelity kernel.
To test whether the conclusion is specific to that choice, we evaluated five quantum models drawn from the Qiskit Machine Learning reference implementations, each against the classical model it is naturally matched to, on identical features and identical folds.

\subsection{Models}

\textbf{QCNN} follows the quantum convolutional network of Cong et al.~\cite{cong2019} as given by the Qiskit tutorial: a ZFeatureMap encoding, alternating convolution and pooling layers, and a Pauli-$Z$ read-out on the surviving qubit. We re-implemented it in a batched statevector simulator
because Qiskit's \texttt{EstimatorQNN} evaluates one sample per circuit execution, which does not scale to $2{,}428$ subjects across $25$ folds; equivalence to the reference is verified to $10^{-6}$.

\textbf{VQC} is the standard variational quantum classifier: ZZFeatureMap encoding \cite{Havlicek2019QuantumFeature}, a RealAmplitudes ansatz, and a parity read-out mapping the measured basis distribution onto two classes. Unlike the ring encoder of Section~\ref{sec:architecture}, which encodes the input as a rotation angle and returns $n$ expectation values, it encodes the input as a phase, once per qubit and again per entangled pair, and collapses the $2^{n}$-dimensional register onto a single parity bit. A phase of $2x$ is $2\pi$-periodic, so a feature scaled to $[0,\pi]$ maps its two extremes onto the same state; both variational arms are therefore scaled to $[0,\pi/2]$, which keeps every feature value on a distinct state.

\textbf{QSVM-fixed} is the fidelity-kernel quantum support vector machine (QSVM) \cite{gentinetta2024} of Section~\ref{sec:matched-comparison}, with bandwidth selected by kernel-target alignment. \textbf{QSVM-Pegasos} applies Qiskit's \texttt{PegasosQSVC} to the same precomputed kernel.

\textbf{TQEK} is a trainable quantum embedding kernel~\cite{hubregtsen2022}, in which the embedding's parameters are optimised against kernel-target alignment so the feature geometry adapts to the task. Section~\ref{sec:kernel-spectrum} motivates it directly: a fixed kernel has one free parameter in its width and therefore traverses the same family of Gram matrices as an RBF kernel, whereas a trainable embedding does not.

Each quantum model is paired with a classical comparator of matched capacity and identical training budget. TQEK's comparator is an RBF kernel on a \emph{learned} linear metric, optimised under the same objective, optimiser and step count. Without it, a trained quantum kernel beating an untrained
classical one would demonstrate only that training helps.

\subsection{Results}

\begin{figure}[!tbp]
\centering
\includegraphics[width=\columnwidth]{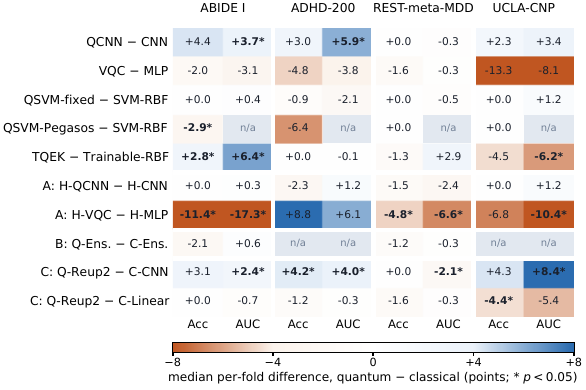}
\caption{Median per-fold accuracy and AUC difference between each quantum model and its matched classical twin, for the model suite and the three interventions.}
\label{fig:ledger}
\end{figure}

Table~\ref{tab:comprehensive} reports every model on every cohort across five metrics. Quantum models, marked $\dagger$, and their matched comparators see identical features and identical folds; SVM rows labelled ``2,000 feat.'' receive the full $2{,}000$ supervised-selected features, the population-graph rows (quantum population graph, QPG; classical population graph, PG; and the no-graph MLP) use the $128$-dimensional descriptors of
Section~\ref{sec:multicohort}, and \texttt{PegasosQSVC} reports no AUC.
Fig.~\ref{fig:ledger} shows the median per-fold difference in accuracy and AUC between every quantum model and its matched twin, for this suite and for the interventions of Section~\ref{sec:interventions}, with $p<0.05$ on a paired Wilcoxon test marked; the differences quoted below are these medians.

Twenty paired accuracy comparisons were made. Two reach $p<0.05$ uncorrected, both on ABIDE~I and in opposite directions: TQEK beats its comparator ($+0.028$, $p=0.049$) while QSVM-Pegasos loses to its own ($-0.029$, $p=0.039$). With twenty tests, one expects roughly one false positive at $\alpha=0.05$; the inconsistent effect directions and the failure of both to survive Bonferroni correction ($0.05/20 = 0.0025$) are compatible with sampling variation.

TQEK is the only quantum model with a significant accuracy win anywhere in this suite, and it does not replicate: $+0.028$ ($p=0.049$) on ABIDE~I but $-0.013$ ($p=0.27$) on REST-meta-MDD and $-0.045$ ($p=0.088$) on UCLA-CNP. Section~\ref{sec:multicohort} shows similar cohort-dependent behaviour. 

QCNN is the most consistent quantum model in the suite: it leads its matched classical convolutional network on ABIDE~I ($+0.044$) and UCLA-CNP ($+0.023$), matches it on REST-meta-MDD, and is also ahead on ADHD-200 ($+0.030$), where that comparator does not separate the classes (Section~\ref{sec:quantum-wins}). The ABIDE~I difference approaches significance ($p=0.058$), indicating a consistent direction of effect, and QCNN is the only quantum model that does not trail its comparator on any cohort.

\texttt{PegasosQSVC} exposes no decision function, so no AUC is available, and on ADHD-200 it predicts a single class on every fold ($\mathrm{F1}=0.000$ at the majority-class accuracy of $0.561$). Runtimes favour the classical side by up to two orders of magnitude: on REST-meta-MDD, QSVM-fixed takes $0.67$\,s per fold against $0.06$\,s for SVM (RBF), TQEK $3.0$\,s against $0.10$\,s for Trainable-RBF, VQC $15.2$\,s against $0.07$\,s for the MLP, and QCNN $32.6$\,s against $8.6$\,s for the classical CNN, on an exact simulator where the quantum models incur no shot noise or device overhead.

\section{Multi-Cohort Validation}
\label{sec:multicohort}

The ABIDE results suggest performance parity under matched feature constraints. To examine whether this observation generalises across disorders and preprocessing pipelines, we applied the same framework to three further public cohorts (Table~\ref{tab:cohorts}). All design choices were fixed on ABIDE~I before these cohorts were analysed, so they provide an independent, out-of-sample evaluation.

\subsection{Quantum Kernel Versus Classical Kernels}

We evaluate the subject-level quantum kernel $K_{ij}=|\langle\psi(x_i)|\psi(x_j)\rangle|^2$ (here $x_i$ is the descriptor of subject $i$) against classical kernels on \emph{identical} descriptors and \emph{identical} folds. Encoding bandwidth and kernel type are selected within each fold by kernel-target alignment,
\begin{equation}
\mathrm{KTA}(K, y) = \frac{\langle K,\, yy^{\top}\rangle_F}{\|K\|_F\, \|yy^{\top}\|_F},
\label{eq:kta}
\end{equation}
where $K$ is the Gram matrix of the $m$ training subjects, $y \in \{-1,+1\}^{m}$ holds their labels, and $\langle\cdot,\cdot\rangle_F$ and $\|\cdot\|_F$ are the Frobenius inner product and norm. The classical comparator additionally optimises its width parameter, providing at least comparable tuning flexibility. Because both models see the same folds, significance is assessed with a paired Wilcoxon signed-rank test over per-fold scores.

A second framing places subjects rather than regions on the graph. In this population graph, each node is a subject carrying a $128$-dimensional connectivity descriptor, edges come from the eight-qubit fidelity kernel (QPG) or from classical RBF or correlation kernels (PG), and an identical GCN
classifies every variant; an MLP given the same descriptors with no graph is the reference. On ABIDE~I, the quantum edges reach an accuracy of $0.537$, against $0.589$ and $0.601$ for correlation and RBF edges and $0.662$ for the no-graph MLP.

\subsection{Evaluation of Replication Across Cohorts}

On UCLA-CNP, the quantum kernel outperforms the classical linear kernel on $8$ of $10$ folds ($p=0.037$), whereas on REST-meta-MDD the same comparison favours the linear kernel, which wins on $18$ of $25$ folds ($p=0.011$).

REST-meta-MDD is the stronger evidence, with roughly ten times as many subjects and folds that hold out whole sites, so the UCLA-CNP result more likely reflects cohort-specific variability than a reproducible effect. Three further observations support this reading: on UCLA-CNP the AUC of every model in this comparison sits at chance ($0.490$--$0.514$); the advantage appears only against the linear kernel and never against the RBF kernel on any cohort; and of the six paired tests across the three further cohorts, the two that reached $p<0.05$ point in opposite directions and neither survives Bonferroni correction ($0.05/6=0.008$).

On REST-meta-MDD the full-dimensional classical RBF reaches $\mathrm{F1}=0.671$ and $\mathrm{AUC}=0.684$, against $0.571$ and $0.544$ for the quantum kernel on eight principal components. 
\section{Limits of the Fidelity Graph}\label{sec:graph-limits}
This section examines two properties of fidelity-based graph construction that hold beyond our implementation (the register width at which fidelity still supplies usable edges, and an algebraic invariance that fixes which circuit parameters enter it) and then reports the graph pipeline's results.

\subsection{Register Width Determines Whether Fidelity Is Usable}
\label{sec:fidelity-collapse}

\begin{figure}[!tbp]
\centering
\includegraphics[width=\columnwidth]{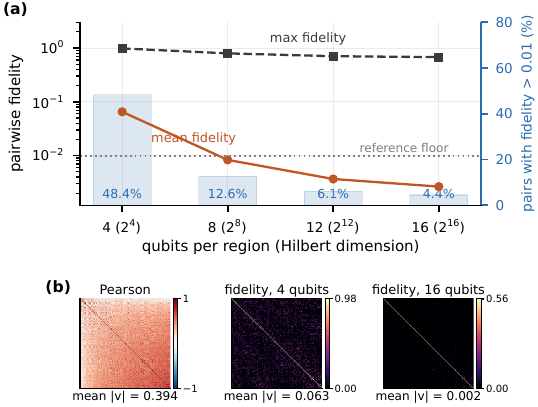}
\caption{Fidelity concentration with register width over twenty ABIDE~I subjects, and one subject's connectivity under Pearson correlation and four- and sixteen-qubit fidelity.}
\label{fig:fidelity-collapse}
\end{figure}

Fidelity remains useful for graph construction only while sufficient numbers of region pairs exhibit measurable overlap. Measuring all $19{,}900$ region pairs on twenty subjects, with each region described by the connectivity profile the graph pipeline uses, shows how rapidly pairwise fidelity decreases with increasing register width. Fig.~\ref{fig:fidelity-collapse}(a) plots the mean and maximum pairwise fidelity on a log scale against a reference floor of $0.01$, with the share of region pairs above that floor as bars. From four qubits to eight, twelve and sixteen, the Hilbert dimension grows from $16$ to $256$, $4{,}096$ and $65{,}536$; mean fidelity falls from $0.0652$ to $0.0083$, $0.0037$ and $0.0027$, crossing the floor between four and eight qubits; and the share of pairs above the floor falls from $48.4\%$ to $12.6\%$, $6.1\%$ and $4.4\%$.

What matters is not the share but whether enough pairs survive to build the graph. A $k=20$ graph over $200$ regions needs $4{,}000$ directed edges, spread over at least $2{,}000$ distinct pairs. At eight qubits, a median of $2{,}336$ pairs clear the floor, but unevenly, so $11\%$ of a fidelity-only graph's edges already fall below it; at twelve and sixteen qubits only $1{,}108$ and $787$ pairs clear it, and $48\%$ and $61\%$ of the edges are taken from below-floor similarities. The floor is a convention, and the crossover moves with it: at $0.003$ twelve qubits still supply $2{,}246$ pairs, at $0.001$ even sixteen supply $2{,}853$, and at $0.03$ eight qubits supply only $1{,}169$. What holds at every floor is the direction: each added qubit leaves fewer pairs, and a smaller share of the graph's edges, above it. The parcellation supplies 200 regional states; a $65{,}536$-dimensional space has far more room than 200 states can populate, so the typical overlap falls toward zero. This is the concentration effect established for quantum kernels \cite{Thanasilp2024Concentration}, though on structured connectivity features it is far slower than for random states: mean fidelity at sixteen qubits is $0.0027$, about $180$ times the $2^{-16} \approx 1.5\times10^{-5}$ of random states. It shares a root with barren plateaus \cite{McClean2018Barren, Cerezo2021BarrenPlateaus}, namely measure concentration in an exponentially large Hilbert space, but the two are distinct: concentration makes kernel \emph{values} indistinguishable, and bites even where no gradient is taken, whereas a barren plateau flattens a variational \emph{gradient}. Section~\ref{sec:width} separates them here.

This concentration limits the usefulness of wider registers for the evaluated fidelity graph. A graph remains usable under near-orthogonality only while enough region pairs stay non-orthogonal; once the bulk of the matrix is orthogonal, too few pairs are left to fill a $k=20$ neighbourhood, and the selection step must draw an increasing proportion of edges from below-floor similarities, whose reliability as connectivity estimates has not been established. On representative subjects, the Pearson matrix is dense (average degree $k \approx 80$) and four-qubit fidelity markedly sparser ($k \approx 25$ after thresholding); for one subject the mean absolute value falls from $0.394$ under Pearson correlation to $0.063$ at four qubits and $0.002$ at sixteen, where the matrix is sparse apart from its diagonal
(Fig.~\ref{fig:fidelity-collapse}(b)). The practical consequence is an operating rule, not a bound. Its general half is arithmetic: a $k$-nearest-neighbour graph over $N$ regions needs $kN$ edges on at least $kN/2$ distinct pairs, and each region needs $k$ above-floor neighbours of its own, so the pair count is necessary but not sufficient. Its other half is empirical, since the crossover depends on the encoder, the parcellation, the feature parameterisation and the floor itself; on these data, at $N=200$ and $k=20$, the widest register that still supplies enough pairs falls from sixteen qubits at a floor of $0.001$ to four at $0.03$, far short of one qubit per feature. We read $2^n \not\gg N$ as a summary of that measurement, not a derived limit.

\subsection{Register Width, Accuracy and Computational Cost in Variational Models}
\label{sec:width}

Section~\ref{sec:fidelity-collapse} concerns \emph{kernels}, which compare states pairwise. A QCNN reads a single Pauli-$Z$ expectation instead, so concentration need not bite the same way, and whether width helps a variational model was an open question rather than a measured one.

Training a QCNN at four, eight and sixteen qubits on a $300$-subject pool with a held-out site gives test accuracies of $0.491$, $0.503$ and $0.520$, with the overfit gap narrowing from $+0.149$ to $+0.110$. The present experiment suggests that fidelity concentration does \emph{not} directly translate into degraded QCNN performance. However, the observed performance gain was achieved at substantial computational cost: sixteen qubits cost $163 \times$ the runtime of eight ($685.5$\,s against $4.2$\,s) for $+0.017$ accuracy, and statevector memory doubles per qubit.

Training accuracy never exceeds $0.65$ at any width, so the model underfits while simultaneously overfitting by $0.11$ to $0.15$. This is a single-site diagnostic rather than a cross-validated result, and no accuracy here should be read as a cohort figure; Section~\ref{sec:plan-c} tests the remaining option, adding capacity at fixed width.

\subsection{The Variational Layer Is Invisible to the Graph}
\label{sec:invariance}

A third limitation is algebraic rather than empirical: it constrains what the fidelity graph can learn at any register width.

Our encoder prepares $|\psi(x)\rangle = U\,V(x)|0\rangle$, where $V(x)$ applies the trainable encoding gains and $U$ is the entangling ring followed by the variational rotations. $U$ is unitary and applied identically to every region, so it cancels exactly in the fidelity, because $U^{\dagger}U = I$:
\begin{equation}
F_{ij} = \bigl|\langle 0|V(x_i)^{\dagger} U^{\dagger} U V(x_j)|0\rangle\bigr|^{2}
= \bigl|\langle 0|V(x_i)^{\dagger} V(x_j)|0\rangle\bigr|^{2}.
\end{equation}

Everything after the data-dependent encoding leaves the fidelity kernel invariant and receives no gradient from the graph. We confirmed this numerically: perturbing the variational parameters by $O(1)$ changes the fidelity Gram matrix by $1.2\times10^{-6}$, and their gradients through the
graph are $1.0\times10^{-7}$ and $3.8\times10^{-8}$ against $51.5$ for the encoding gains. Through the Pauli-$Z$ read-out path the same parameters receive gradients of $0.128$ and $0.298$, so the circuit is not inert. It is only its contribution to the \emph{graph} that vanishes.

The encoding gains are therefore the only circuit parameters that enter a fidelity graph: eight parameters for a $200$-node topology, rather than the full variational stack the architecture appears to provide. In the principal graph runs, the encoder is evaluated once, without gradients, at its initial parameters, and the top-$k$ support selected from that encoding stays fixed
while the edge kernel and the classifier are optimised; training the encoder as well, with the support re-selected every epoch, leaves the graph at chance (Section~\ref{sec:graph-failure}). The invariance therefore does not affect any reported result; it constrains how a fidelity graph can be trained.

The argument extends to any architecture in which identical post-encoding unitary operations are applied before fidelity is computed. In such constructions, only trainable parameters that precede or are interleaved with the data-dependent encoding can influence the fidelity; a unitary block placed after the final encoding layer and applied identically to every input cancels regardless of its depth. Even when every parameter enters the fidelity, the support is fixed by a top-$k$ selection that carries no gradient, so a fidelity graph is trainable in its edge weights but not in its topology.

The identity reaches beyond graphs to any hybrid model that compares data by state overlap, including fidelity kernels in support vector machines and kernel-alignment training whose optimised block follows the encoding. In each case, the kernel's gradient with respect to those layers is exactly zero at every parameter setting, unlike the exponentially small gradients of barren plateaus
\cite{McClean2018Barren, Cerezo2021BarrenPlateaus}, so barren-plateau mitigations such as tailored initialisation \cite{grant2019init} or layer-wise training \cite{skolik2021layerwise} cannot recover it, while the cancelled layers still cost depth, simulation time and, on hardware, gate noise. Trainable similarity therefore has to sit where it survives: inside the feature map, as in trainable embedding kernels~\cite{hubregtsen2022}; between repeated encodings, as in data
re-uploading~\cite{perezsalinas2020}, where only the last trainable block cancels; or in classical layers applied after the overlap, such as graph attention. Perturbing the post-encoding parameters and checking whether the Gram matrix moves is enough to tell which case a published hybrid falls into.

\subsection{Performance Limitations of the Fidelity-Graph Pipeline}
\label{sec:graph-failure}

The graph-attention pipeline stays at chance in every configuration we tried, with accuracies of $0.504$ for the fidelity graph with attention, $0.512$ for its GCN ablation and $0.513$ with a trained encoder, and in almost every fold it assigns the whole held-out site to one class (MCC $0.000$ for both fidelity-graph models; Table~\ref{tab:results-graph}). Modifying the kernel, changing the read-out, supplying node features that a linear classifier separates at AUC $0.739$, and training the encoder jointly while re-selecting the support every epoch did not materially improve performance, and a quantum graph transformer behaved the same way. Because informative node features did not help, the remaining candidate explanations include sparsification effects, the quality of the fidelity-selected topology, limitations of attention learning and graph over-smoothing; the present study does not isolate their individual contributions, so we report the graph arm as a negative result, and every quantum--classical comparison in this paper rests on the feature-based models.

\begin{table}[!b]
\caption{Every model on every cohort. All values are means over folds; leave-site-out except UCLA-CNP, which is stratified 10-fold. $\dagger$ marks a quantum model. SVM labels give each model's input: eight matched features, or the 2,000 supervised-selected ones.}
\label{tab:comprehensive}
\centering
\scriptsize
\setlength{\tabcolsep}{1.6pt}
\renewcommand{\arraystretch}{0.95}
\begin{tabular}{|l|c|c|c|c|c|}
\hline
\textbf{Model} & \textbf{Acc} & \textbf{F1} & \textbf{AUC} & \textbf{Spec} & \textbf{MCC} \\
\hline
\multicolumn{6}{|l|}{\textit{ABIDE~I}} \\
\hline
QCNN$\dagger$ & $0.615$ & $0.561$ & $0.666$ & $0.699$ & $0.253$ \\
QPG (quantum edges)$\dagger$ & $0.537$ & $0.528$ & $0.576$ & $0.514$ & $0.074$ \\
QSVM-Pegasos$\dagger$ & $0.579$ & $0.475$ & --- & $0.750$ & $0.199$ \\
QSVM-fixed$\dagger$ & $0.622$ & $0.560$ & $0.663$ & $0.717$ & $0.264$ \\
TQEK$\dagger$ & $0.610$ & $0.534$ & $0.686$ & $0.724$ & $0.234$ \\
VQC$\dagger$ & $0.582$ & $0.535$ & $0.612$ & $0.632$ & $0.179$ \\
CNN (1-D) & $0.589$ & $0.538$ & $0.638$ & $0.652$ & $0.194$ \\
MLP & $0.600$ & $0.550$ & $0.629$ & $0.674$ & $0.208$ \\
MLP (no graph) & $0.662$ & $0.641$ & $0.729$ & $0.688$ & $0.335$ \\
PG (RBF edges) & $0.601$ & $0.589$ & $0.630$ & $0.593$ & $0.211$ \\
PG (correlation edges) & $0.589$ & $0.587$ & $0.639$ & $0.574$ & $0.195$ \\
SVM (RBF), 8 feat. & $0.616$ & $0.559$ & $0.662$ & $0.702$ & $0.255$ \\
SVM (RBF), 2,000 feat. & $0.649$ & $0.632$ & $0.738$ & $0.642$ & $0.323$ \\
SVM (Linear), 2,000 feat. & $0.629$ & $0.624$ & $0.692$ & $0.618$ & $0.268$ \\
Trainable-RBF & $0.581$ & $0.562$ & $0.606$ & $0.586$ & $0.170$ \\
\hline
\multicolumn{6}{|l|}{\textit{ADHD-200}} \\
\hline
QCNN$\dagger$ & $0.512$ & $0.408$ & $0.551$ & $0.609$ & $0.019$ \\
QPG (quantum edges)$\dagger$ & $0.526$ & $0.363$ & $0.558$ & $0.694$ & $0.026$ \\
QSVM-Pegasos$\dagger$ & $0.561$ & $0.000$ & --- & $0.998$ & $-0.009$ \\
QSVM-fixed$\dagger$ & $0.546$ & $0.458$ & $0.561$ & $0.625$ & $0.105$ \\
TQEK$\dagger$ & $0.549$ & $0.421$ & $0.551$ & $0.674$ & $0.074$ \\
VQC$\dagger$ & $0.503$ & $0.426$ & $0.538$ & $0.544$ & $0.015$ \\
CNN (1-D) & $0.495$ & $0.385$ & $0.504$ & $0.609$ & $0.018$ \\
MLP & $0.540$ & $0.437$ & $0.548$ & $0.619$ & $0.060$ \\
MLP (no graph) & $0.535$ & $0.234$ & $0.522$ & $0.824$ & $-0.014$ \\
PG (RBF edges) & $0.524$ & $0.308$ & $0.493$ & $0.730$ & $-0.001$ \\
PG (correlation edges) & $0.503$ & $0.285$ & $0.495$ & $0.726$ & $-0.013$ \\
SVM (RBF), 8 feat. & $0.571$ & $0.448$ & $0.573$ & $0.683$ & $0.125$ \\
SVM (RBF), 2,000 feat. & $0.561$ & $0.264$ & $0.555$ & $0.841$ & $0.034$ \\
SVM (Linear), 2,000 feat. & $0.508$ & $0.331$ & $0.489$ & $0.702$ & $0.006$ \\
Trainable-RBF & $0.562$ & $0.430$ & $0.554$ & $0.688$ & $0.094$ \\
\hline
\multicolumn{6}{|l|}{\textit{REST-meta-MDD}} \\
\hline
QCNN$\dagger$ & $0.517$ & $0.500$ & $0.561$ & $0.581$ & $0.054$ \\
QPG (quantum edges)$\dagger$ & $0.526$ & $0.528$ & $0.550$ & $0.557$ & $0.059$ \\
QSVM-Pegasos$\dagger$ & $0.523$ & $0.508$ & --- & $0.573$ & $0.074$ \\
QSVM-fixed$\dagger$ & $0.522$ & $0.513$ & $0.547$ & $0.577$ & $0.060$ \\
TQEK$\dagger$ & $0.519$ & $0.454$ & $0.570$ & $0.665$ & $0.071$ \\
VQC$\dagger$ & $0.505$ & $0.469$ & $0.540$ & $0.599$ & $0.029$ \\
CNN (1-D) & $0.540$ & $0.514$ & $0.568$ & $0.619$ & $0.102$ \\
MLP & $0.523$ & $0.496$ & $0.553$ & $0.599$ & $0.059$ \\
MLP (no graph) & $0.608$ & $0.632$ & $0.652$ & $0.578$ & $0.220$ \\
PG (RBF edges) & $0.573$ & $0.612$ & $0.601$ & $0.494$ & $0.139$ \\
PG (correlation edges) & $0.585$ & $0.610$ & $0.602$ & $0.545$ & $0.158$ \\
SVM (RBF), 8 feat. & $0.526$ & $0.499$ & $0.561$ & $0.603$ & $0.076$ \\
SVM (RBF), 2,000 feat. & $0.590$ & $0.604$ & $0.651$ & $0.567$ & $0.182$ \\
SVM (Linear), 2,000 feat. & $0.576$ & $0.593$ & $0.594$ & $0.562$ & $0.152$ \\
Trainable-RBF & $0.571$ & $0.682$ & $0.566$ & $0.238$ & $0.104$ \\
\hline
\multicolumn{6}{|l|}{\textit{UCLA-CNP}} \\
\hline
QCNN$\dagger$ & $0.585$ & $0.610$ & $0.589$ & $0.566$ & $0.168$ \\
QSVM-Pegasos$\dagger$ & $0.561$ & $0.652$ & --- & $0.324$ & $0.094$ \\
QSVM-fixed$\dagger$ & $0.554$ & $0.584$ & $0.579$ & $0.519$ & $0.101$ \\
TQEK$\dagger$ & $0.558$ & $0.594$ & $0.569$ & $0.507$ & $0.111$ \\
VQC$\dagger$ & $0.486$ & $0.509$ & $0.528$ & $0.485$ & $-0.030$ \\
CNN (1-D) & $0.544$ & $0.569$ & $0.555$ & $0.518$ & $0.093$ \\
MLP & $0.597$ & $0.630$ & $0.611$ & $0.539$ & $0.197$ \\
SVM (RBF), 8 feat. & $0.540$ & $0.570$ & $0.569$ & $0.506$ & $0.077$ \\
SVM (RBF), 2,000 feat. & $0.567$ & $0.608$ & $0.591$ & $0.508$ & $0.120$ \\
SVM (Linear), 2,000 feat. & $0.557$ & $0.627$ & $0.629$ & $0.408$ & $0.083$ \\
Trainable-RBF & $0.602$ & $0.634$ & $0.610$ & $0.557$ & $0.201$ \\
\hline
\end{tabular}
\\[2pt]
{\scriptsize $\dagger$ denotes a quantum model.}
\end{table}

\section{Architectural Interventions on the Feature Budget}
\label{sec:interventions}

Section~\ref{sec:matched-comparison} located the deficit in the register's feature budget rather than in the kernel: at eight features the quantum and classical kernels are indistinguishable, while a classical model given all $2{,}000$ features is better. We therefore designed three interventions that lift that budget without widening the register, each reported with matched classical twins receiving the identical construction, because an intervention that helps only when the classical arm is denied it measures the comparison rather than the model. Rows A--C of Fig.~\ref{fig:ledger} summarise the matched comparisons.

\subsection{A Learned Projection}
\label{sec:plan-a}

Fixed $t$-test selection is replaced by a linear map $\mathbb{R}^{2000} \to \mathbb{R}^{8}$ learned jointly with the circuit, so the quantum stage receives an optimised summary of every connection rather than eight raw ones. Every quantum head is given a classical twin with an identical projection, identical bottleneck width and identical training, plus a linear-head control that
measures how much of any gain comes from the projection alone. The prefix H- marks these projection-fed heads.

The projection helps the classical arm and leaves the convolutional quantum head statistically indistinguishable from its classical twins. On ABIDE~I, H-QCNN reaches $0.645$, the best model in the projection comparison, against $0.641$ for H-CNN and $0.630$ for H-Linear; neither comparison approaches significance ($p=0.37$ and $p=0.18$). On REST-meta-MDD it reaches $0.573$ against $0.589$ and $0.591$, again with no significant separation ($p=0.27$, $p=0.25$).

\textbf{The variational classifier is a different matter.} H-VQC did not improve, and its deficit is the strongest effect in the study. Across four cohorts, nine of forty pairwise comparisons reach $p<0.05$ against an expectation of two, three survive Bonferroni correction, and \emph{eight of the nine have H-VQC on the losing side}. On REST-meta-MDD it loses to H-Linear by $0.049$ ($p=0.004$) and to H-MLP by $0.048$ ($p=0.048$). The evaluated configuration, a ZZ feature map with a RealAmplitudes ansatz and a parity read-out, performed worse than the corresponding classical heads, whereas the convolutional quantum head with a Pauli-$Z$ read-out did not. H-Linear reaches within $0.02$ of the best classical head on every cohort, so the projection carries most of the signal; additional read-out ablations are required to determine which component of the variational configuration causes the deficit.

\subsection{Ensembles Over Disjoint Feature Blocks}
\label{sec:plan-b}

Where the projection compresses all features into the register, ensembling sidesteps the cap: the top $32$ features are split into four disjoint blocks, one narrow model is trained per block, and the outputs are combined with learned weights.
Every circuit stays at eight qubits while the ensemble sees $32$ features. Two comparators separate the two things that could produce a gain. A classical ensemble over the identical blocks with the identical combiner isolates the model; a single classical model given all $32$ features at once bounds what the partition itself costs.

Ensembling helps the classical arm and the quantum arm equally. On ABIDE~I the block ensemble reaches $0.618$ for the quantum members against $0.622$ for the matched classical ensemble, and on REST-meta-MDD $0.540$ against $0.542$; of six pairwise comparisons across these two cohorts, none reaches $p<0.05$ (smallest $p=0.178$). On ABIDE~I, both ensembles exceed the single classical model given all $32$ features at once ($0.603$), so the partition is worth roughly $+0.017$ whether the members are quantum or classical, but on REST-meta-MDD the single model leads ($0.564$), so even the ensembling benefit does not replicate.

\subsection{Data Re-uploading}
\label{sec:plan-c}

The underfitting observed in Section~\ref{sec:width} points to capacity rather than width. Data re-uploading~\cite{perezsalinas2020} re-injects the features between convolution stages, each time with its own trainable gain, raising the order of Fourier terms the circuit can express, a capacity that a single input encoding cannot reach, at an unchanged register width. We compare a plain QCNN against re-upload depths one and two, alongside the matched classical heads C-CNN, C-MLP and C-Linear, and record \emph{training} accuracy as well as test accuracy, because the claim being tested is specifically about capacity.

Re-uploading does not improve on the plain QCNN on any cohort. The median per-fold change at depth one is $0.000$ on ABIDE~I, ADHD-200 and REST-meta-MDD ($p \ge 0.63$) and $-0.043$ on UCLA-CNP ($p=0.11$), and depth two is no better ($p \ge 0.09$ throughout). Across all four cohorts, $60$ pairwise comparisons among the six models yield nine results at $p<0.05$ uncorrected against an
expectation of three by chance, and \emph{none} survives Bonferroni correction ($0.05/60 = 0.00083$; the smallest $p$ anywhere is $0.008$).

All nine sit on the two smallest cohorts, and seven of them have the same losing side: C-CNN. On ADHD-200, all three quantum models beat it (medians $+0.038$ to $+0.048$, $p=0.016$ to $0.031$), but so does C-Linear ($+0.062$, $p=0.031$); on UCLA-CNP the plain QCNN beats it ($+0.087$, $p=0.008$), and so do C-MLP ($+0.087$, $p=0.014$) and C-Linear ($+0.111$, $p=0.014$), at the same effect size. The remaining two go against the quantum arm, with C-Linear beating both re-uploading circuits on UCLA-CNP ($+0.045$ and $+0.044$, $p=0.031$). The 1-D CNN fails to train at these sample sizes, sitting at or below chance ($0.482$ on ADHD-200, $0.500$ on UCLA-CNP) while reaching $0.599$ on ABIDE~I.

\subsection{Capacity Converts to Overfitting, Not Accuracy}
\label{sec:capacity}

Recording training accuracy clarifies this null result: across the evaluated cohorts, models with higher training accuracy did not achieve higher held-out accuracy. Across the six re-uploading models, mean training accuracy predicts the overfit gap with a Pearson correlation of $r=+0.983$ ($p=0.0004$) on REST-meta-MDD, $+0.949$ ($p=0.004$) on ABIDE~I, $+0.731$ ($p=0.099$) on ADHD-200 and $+0.612$ ($p=0.197$) on UCLA-CNP, positive on all four and significant on the two best-powered cohorts. Parameter count tracks the gap even more consistently (Spearman $\rho = +0.94$, $+0.94$, $+0.89$, $+1.00$; significant on all four). Training accuracy is meanwhile \emph{negatively} related to test accuracy on three cohorts.

Re-uploading did what its theory claims: it raised training accuracy from $0.651$ to $0.655$ on ABIDE~I and from $0.575$ to $0.578$ on REST-meta-MDD, so the added Fourier capacity is real and usable; it did not, however, improve held-out accuracy. On ABIDE~I the best re-uploading model is an $18$-parameter linear layer, with the best accuracy, the best AUC, the smallest gap and a runtime two orders of
magnitude below the QCNN. On REST-meta-MDD, all six models sit inside a $0.008$ band.

This revises the reading of Section~\ref{sec:width}: additional capacity did not raise the QCNN's held-out accuracy. Increasing capacity, by widening the register, learning a projection, ensembling, or re-uploading, improved training fit but did not improve held-out performance in the evaluated configurations, whether the mechanism adding it was quantum or classical.

Taken with Sections~\ref{sec:multicohort} and~\ref{sec:quantum-suite}, six architectural families have now been tested: fidelity kernels, population graphs, the Qiskit model suite, learned projections, block ensembles and data re-uploading. In every one, each intervention that helps the quantum arm helps its matched classical twin at least as much.

\section{Discussion: When Does Quantum Help?}
\label{sec:when-quantum}

\subsection{Where the Arms Differ}
\label{sec:regularisation}

Comparing training and test accuracy reveals one aspect in which the model families differ. A QCNN reads out a single Pauli-$Z$ expectation bounded in $[-1,1]$, and the affine map to logits adds two parameters. All information the circuit learns is therefore conveyed through this bounded scalar output. This bottleneck may contribute to the smaller train--test gap observed in the quantum models, but we have not isolated it; the control would be a classical network of matched size ending in the same bounded scalar.

\begin{figure}[!tbp]
\centering
\includegraphics[width=\columnwidth]{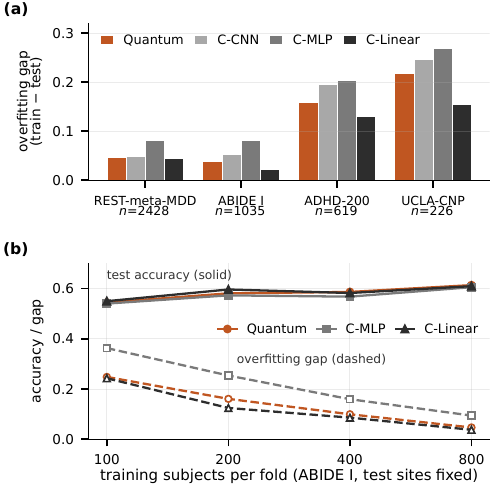}
\caption{Overfitting gap by cohort, and test accuracy and overfitting gap against training-set size on ABIDE~I.}
\label{fig:regularisation}
\end{figure}

The overfitting gap, training minus test accuracy, of the quantum arms averaged over the three QCNN variants of Section~\ref{sec:plan-c} is smaller than that of both neural comparators on all four cohorts (Fig.~\ref{fig:regularisation}(a)), though the linear classifier, the smallest model, overfits least on every cohort, and across all six models the gap tracks parameter count (Section~\ref{sec:capacity}). The difference is therefore consistent with lower effective capacity; it does not isolate a quantum-specific mechanism and does not, by itself, establish a quantum advantage. The margin over the MLP is $0.034$, $0.043$, $0.043$ and $0.051$ for cohorts of $2{,}428$, $1{,}035$, $619$ and $226$ subjects (paired Wilcoxon $p<0.001$, $p=0.001$, $0.047$ and $0.193$). These observations should be interpreted descriptively rather than inferentially: with four cohorts the exact permutation null gives $p=0.333$ for a rank correlation of this size, and the \emph{ratio} of the two gaps orders the cohorts the other way ($1.72$, $2.13$, $1.27$, $1.23$).

\subsection{Analysis of Positive Quantum--Classical Differences}
\label{sec:quantum-wins}

We examined the individual matched comparisons to identify patterns not reflected in the aggregate results. Across all runs, we make $284$ matched comparisons, covering every quantum/classical pair, on accuracy, F1, AUC, and specificity. The quantum arm is ahead in $115$ of them ($40.5\%$) and significantly ahead in $30$ ($10.6\%$; Table~\ref{tab:advantage-sweep}). There, ``ahead'' counts a positive median
difference for the quantum arm, ``sig.'' counts $p<0.05$ on a paired Wilcoxon test, and ``clean'' excludes wins in which either arm is not classifying (sensitivity or specificity below $0.30$, or $|\mathrm{MCC}|<0.02$, on the confusion matrix pooled over folds) or the comparator fails to train (C-CNN).

\begin{table}[!tbp]
\caption{Complete sweep of matched quantum--classical comparisons by metric.}
\label{tab:advantage-sweep}
\centering
\footnotesize
\setlength{\tabcolsep}{5pt}
\renewcommand{\arraystretch}{1.05}
\begin{tabular}{|l|r|r|r|r|}
\hline
\textbf{Metric} & \textbf{Pairs} & \textbf{Ahead} & \textbf{Sig.} & \textbf{Clean} \\
\hline
Accuracy & $72$ & $23$ & $5$ & $1$ \\
F1 & $72$ & $35$ & $4$ & $2$ \\
AUC & $68$ & $34$ & $16$ & $6$ \\
Specificity & $72$ & $23$ & $5$ & $3$ \\
\hline
\textbf{Total} & $\mathbf{284}$ & $\mathbf{115}$ & $\mathbf{30}$ & $\mathbf{12}$ \\
\hline
\end{tabular}
\end{table}

\begin{figure}[!tbp]
\centering
\includegraphics[width=\columnwidth]{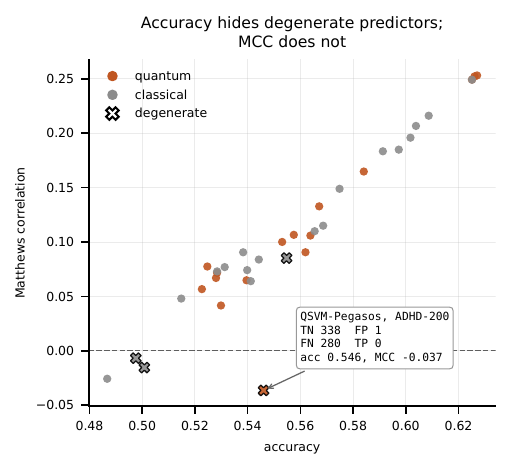}
\caption{Accuracy against Matthews correlation for every model on every cohort, pooled over folds.}
\label{fig:degeneracy}
\end{figure}

\textbf{Eighteen of the thirty are screened out}
(Fig.~\ref{fig:degeneracy}). Plotting accuracy against Matthews correlation for every model on every cohort, pooled over folds, separates the models that classify from those that do not: models predicting close to a single class sit far below the trend, scoring near the majority-class rate on accuracy and near zero on MCC, and the annotated case, QSVM-Pegasos on ADHD-200, returns zero true positives. Screening the classical arms on these pooled matrices flags four that are not classifying: C-CNN on UCLA-CNP ($\mathrm{MCC}=-0.005$), the suite's 1-D CNN ($-0.015$) and C-MLP ($+0.015$) on ADHD-200, and Trainable-RBF on REST-meta-MDD, which is near-all-positive (specificity $0.276$). Fifteen significant wins are against C-CNN; one is a QSVM-Pegasos specificity win earned by predicting a single class; one is TQEK's $+0.471$ specificity against the degenerate Trainable-RBF; and one is the QCNN's ADHD-200 AUC win against the degenerate 1-D CNN. These cases do not provide strong evidence regarding quantum-specific behaviour.

\textbf{Twelve survive, and half of them are on AUC, a ranking metric.}

\begin{itemize}
\item \textbf{QCNN against its matched classical CNN on ABIDE~I.} AUC $+0.037$ ($p=0.033$), specificity $+0.038$ ($p=0.030$), and accuracy $+0.044$ at $p=0.058$. The threshold-independent AUC lead extends to ADHD-200 ($+0.059$, $p=0.047$), reported separately as that comparator does not separate the classes.
\item \textbf{TQEK against a trainable classical kernel on ABIDE~I.} AUC $+0.064$ ($p=0.0012$), specificity $+0.111$ ($p=0.0006$), accuracy $+0.028$ ($p=0.049$). This represents the largest observed single-cohort effect, against a comparator that is functional in this cohort. However, the effect was not reproduced in the remaining cohorts ($-0.013$ on REST-meta-MDD, $-0.045$ on UCLA-CNP).
\item \textbf{Re-uploading against C-MLP on ABIDE~I.} AUC $+0.055$ ($p=0.0094$), $+0.043$ ($p=0.019$) and $+0.053$ ($p=0.0032$) at depths zero, one and two.
\item \textbf{Learned-projection quantum head against its linear control.} AUC $+0.017$ ($p=0.024$) on ABIDE~I.
\end{itemize}

\textbf{Matched versus strongest classical comparators.}
These results refer to matched comparators rather than to the highest-performing classical model: each quantum model is set against the classical model it corresponds to, which we regard as the fairest test of the architecture. Under this comparison, the QCNN achieved higher AUC than its matched 1-D CNN on ABIDE~I, although this difference does not establish superiority over the strongest classical models. Against the best classical AUC in each run, the quantum arm ties on ABIDE~I ($0.679$ against $0.679$ for the linear classifier) and trails on the other three cohorts ($0.548$ against $0.563$; $0.566$ against $0.582$; $0.590$ against $0.603$), so, in general, the quantum models did not outperform the strongest classical alternatives.

\textbf{Why AUC and not accuracy.}\label{sec:calibration} The quantum arm leads on AUC in $34$ of $68$ comparisons but on accuracy in only $23$ of $72$. AUC measures ranking; accuracy measures ranking \emph{and} where the decision threshold falls, and a bounded Pauli-$Z$ expectation mapped to logits by a two-parameter affine layer is a potentially limited calibration mechanism, suggesting threshold calibration as a possible explanation. We tested it on all four cohorts, choosing the threshold that maximises Youden's $J$ (sensitivity plus specificity minus one) on the \emph{training} scores of each fold and applying it unchanged to the test scores, identically for every arm. No arm gains: averaged over the cohorts, calibration changes accuracy by $-0.000$ for the quantum arm, $+0.009$ for C-CNN, $+0.002$ for C-MLP and $-0.009$ for C-Linear, and by $-0.001$ on ABIDE~I, where the quantum arm has its highest AUC ($0.679$). Training-set optimisation of Youden's $J$ therefore did not explain the AUC-to-accuracy discrepancy: the quantum arm leads C-CNN on both ranking and thresholding, and is on par with the linear classifier on both.

\subsection{Assessing Evidence for Quantum Advantage}
\label{sec:bounds}

Three observations limit the evidence for a quantum advantage. The quantum arm is ahead in only $115$ of $284$ matched comparisons ($40.5\%$), where a model with no systematic advantage would be ahead in roughly half. Thirty significant results exceed the fourteen expected by chance, but the four metrics are computed on identical folds and are strongly correlated, so the excess cannot be read as sixteen real effects, and none survives Bonferroni correction across the $284$ comparisons. Finally, the leads do not replicate: almost every clean win is on ABIDE~I, the cohort exhibiting the strongest overall classification performance (best classical accuracy exceeds the majority-class rate by $0.104$, against $0.036$ for REST-meta-MDD and $0.010$ for ADHD-200), and TQEK's result, the strongest in the study, reverses on both other cohorts where it can be measured.

\begin{figure}[!tbp]
\centering
\includegraphics[width=\columnwidth]{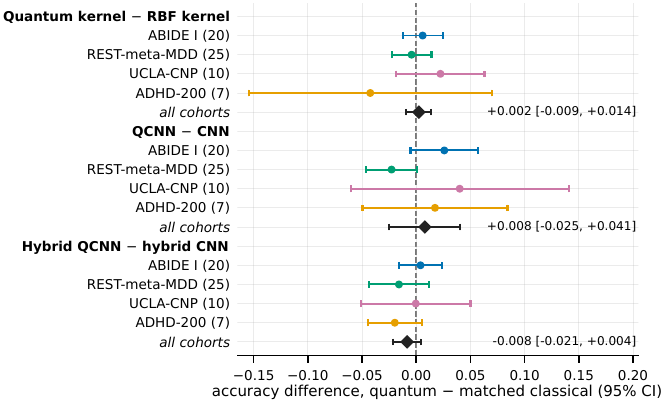}
\caption{Quantum-minus-classical accuracy for the three central matched pairs, per cohort (site folds in brackets) and combined across cohorts by a random-effects model.}
\label{fig:bounds}
\end{figure}

Non-significance is not equivalence, so Fig.~\ref{fig:bounds} bounds the three central matched differences. Combined across cohorts with a random-effects model \cite{dersimonian1986}, the quantum kernel differs from its matched RBF kernel by $+0.002$ in accuracy ($95\%$ CI $-0.009$ to $+0.014$), the QCNN from its CNN by $+0.008$ ($-0.025$ to $+0.041$), and the hybrid QCNN from its hybrid CNN by $-0.008$
($-0.021$ to $+0.004$). Under the fitted random-effects model, the $95\%$ confidence intervals are therefore inconsistent with average quantum improvements larger than $1.4$, $4.1$ and $0.4$ percentage points for the three respective comparisons. ADHD-200, with seven folds, constrains little, so these bounds rest mainly on ABIDE~I and REST-meta-MDD.

A more conservative conclusion is that, on the cohort with the strongest overall classification performance, a quantum convolutional network ranks subjects at least as well as its matched classical counterpart, on a threshold-free metric, with significance. This finding supports competitiveness rather than a reproducible advantage, and it rests on one cohort.

\subsection{The Low-Data Regime, Tested Directly}
\label{sec:lowdata}

Section~\ref{sec:regularisation} left a falsifiable prediction: if the quantum arm's smaller overfitting gap widens as data become scarce, there should be a training size below which it converts into an accuracy advantage. We tested it on ABIDE~I by holding the leave-site-out folds and test sets fixed and subsampling only the training pool to $m$ subjects, three independent draws per fold and size.

Fig.~\ref{fig:regularisation}(b) shows that the gap separates the arms at every size while accuracy does not.

\textbf{The smaller gap holds at every size.} The quantum arm has a smaller overfitting gap than the matched MLP at all four training sizes, without exception, falling from $0.248$ at $m=100$ to $0.160$, $0.099$ and $0.046$ at $200$, $400$ and $800$, against $0.363$, $0.253$, $0.159$ and $0.094$: the MLP overfits $1.46\times$ as much at $m=100$, $1.58\times$ at $200$, $1.61\times$
at $400$ and $2.02\times$ at $800$. The \emph{ratio} grows with sample size rather than shrinking with it: the quantum model maintains a consistently smaller train--test gap, but this relative difference does not increase in the smallest-sample conditions.

\textbf{The direction of the accuracy effect is the opposite of the prediction.} The quantum arm is the best of the three at $m=400$ ($0.586$) and $m=800$ ($0.613$), and \emph{not} at the two smallest sizes, where the linear classifier leads ($0.550$ at $m=100$, $0.596$ at $m=200$). Only one comparison reaches significance, and it favours quantum: $+0.010$ over the MLP at $m=400$
($p=0.040$). A low-data quantum advantage would have produced the reverse ordering.

\subsection{Sources of Apparent Quantum Advantage}
\label{sec:artefacts}

Four mechanisms can produce an apparent quantum advantage, and the matched controls of this study are designed to remove them. Each is common in the published literature, and evaluation defects of this kind have been documented in previous studies: a recent audit of a hybrid quantum--classical skin-lesion classifier found $22.8\%$ of lesions on both sides of the train--test split, inflating every previously reported metric~\cite{jamaludin2026integrity}.

\textbf{An unmatched feature budget.} Comparing an eight-qubit model against a classical model given $2{,}000$ features measures dimensionality, not the kernel; an unmatched configuration shows a quantum deficit for exactly this reason, and it disappears under matching (Section~\ref{sec:matched-comparison}). The budget itself is costly: on REST-meta-MDD, compressing $19{,}900$ pairwise correlations to eight components costs more than any quantum--classical difference measured in this study, while widening the register to admit more features increases the concentration of pairwise fidelity (Section~\ref{sec:fidelity-collapse}).

\textbf{Reliance on a single cohort.} Positive results on one cohort did not replicate on the others anywhere in this study: TQEK gains $+0.028$ on ABIDE~I but loses on REST-meta-MDD and UCLA-CNP (Section~\ref{sec:quantum-suite}), and intervention results on six-fold subsets did not carry over to the full cohorts.

\textbf{A comparator that fails to train.} On ADHD-200, all three quantum models beat the classical CNN at $p \le 0.031$, which may initially be interpreted as a quantum advantage until one observes that C-Linear beats it by a larger margin: the 1-D CNN sits at chance on the small cohorts (Section~\ref{sec:plan-c}), and a single classical comparator would have made this a positive finding.

\textbf{Unreported hyperparameter selection.} The fidelity kernel spans effective dimensions from near-constant to near-identity as bandwidth varies (Section~\ref{sec:kernel-spectrum}); a bandwidth chosen post hoc can place it almost anywhere, and its default leaves it memorising the training set.

These are the reasons every comparison in this paper carries more than one classical arm, reports all cohorts rather than the favourable ones, and states both uncorrected and corrected significance.

\subsection{Comparison with Published Results}\label{sec:placing}
\label{sec:published-comparison}
The proposed pipeline achieves performance that is competitive in some settings but below state-of-the-art results in others: competitive on ABIDE~I, below the strongest published results on ADHD-200 and REST-meta-MDD (Table~\ref{tab:published-comparison}). This outcome reflects the design goals of the study. The classical baseline is deliberately simple: supervised selection plus a standard classifier with no multimodal fusion, no learned harmonisation, and no architecture search, because its role is to give the quantum comparison a well-tuned reference rather than to maximise benchmark performance. On ABIDE~I, the protocol-matched anchor is Abraham et al., whose pipeline reaches about $0.67$ including prediction on unseen sites \cite{abraham2017}; our linear SVM on the full correlation vector reaches $0.671$, matching that level, while a population GCN reaches $0.70$ under ten-fold cross-validation \cite{parisot2018}, which evaluates a different generalisation setting from leave-site-out validation. On ADHD-200, a model given only phenotypic data achieved the highest accuracy of the $21$ teams in the competition ($0.625$, on a three-class task) \cite{brown2012}, which suggests that phenotypic information may contribute substantially to performance on this benchmark. On REST-meta-MDD, an ensemble graph neural network reaches $0.73$ under leave-site-out validation \cite{mddgnn}, above our $0.608$. On UCLA-CNP, a deep graph convolutional network reaches $0.82$ on schizophrenia versus controls under ten-fold cross-validation \cite{sunil2024gnn}, against our $0.602$, but on a different task and $172$ subjects, so it is a reference point rather than a like-for-like comparison.

\begin{table}[!tbp]
\caption{This work against published results on the same cohorts, with the validation protocol of each.}
\label{tab:published-comparison}
\centering
\footnotesize
\setlength{\tabcolsep}{4pt}
\renewcommand{\arraystretch}{1.05}
\begin{tabular}{|l|>{\raggedright\arraybackslash}p{2.9cm}|c|c|}
\hline
\textbf{Cohort} & \textbf{Method} & \textbf{Protocol} & \textbf{Acc} \\
\hline
 \multirow{3}{*}{ABIDE~I} & Abraham et al.~\cite{abraham2017} & LSO & $\sim$0.67 \\
  & Population GCN~\cite{parisot2018} & 10-fold & 0.70 \\
  & \textbf{This work} (linear SVM, full correlation) & LSO & $\mathbf{0.671}$ \\
\hline
 \multirow{2}{*}{ADHD-200} & Phenotypic data alone, three-class~\cite{brown2012} & held-out & 0.625 \\
  & \textbf{This work} (SVM-RBF) & LSO & $\mathbf{0.571}$ \\
\hline
 \multirow{2}{*}{REST-meta-MDD} & Ensemble GNN~\cite{mddgnn} & LSO & 0.73 \\
  & \textbf{This work} (MLP, no graph) & LSO & $\mathbf{0.608}$ \\
\hline
 \multirow{2}{*}{UCLA-CNP} & Deep graph CNN, schizophrenia vs.\ controls~\cite{sunil2024gnn} & 10-fold & 0.82 \\
  & \textbf{This work} (Trainable-RBF) & 10-fold & $\mathbf{0.602}$ \\
\hline
\end{tabular}
\end{table}

\label{sec:comparability}

Reported accuracies on ABIDE~I span from below $0.70$ under leave-site-out validation to above $0.90$ under random splits, and the spread is protocol rather than method. Three conditions must match before two numbers can be placed side by side: the validation protocol, since random $k$-fold splits let a model exploit site-specific artefacts while leave-site-out forbids it; the modality and covariates, since a model given phenotypic variables is solving a different problem; and the parcellation, since atlas choice changes both node count and signal.

DAGMNet~\cite{Wang2025DAGMNet} illustrates all three at once and is the highest ABIDE~I accuracy we encountered ($0.9159$, AUC $0.9680$). It is ten-fold rather than leave-site-out; it fuses sMRI, fMRI, and phenotype rather than using connectivity alone; it reports on the Harvard-Oxford, Eickhoff-Zilles (EZ) and AAL atlases rather than CC200; and its phenotypic inputs include study location, so site identity is available to the model while subjects from the same site appear in both training and test folds, and may therefore contribute to predictive performance under the reported validation protocol.
\section{Limitations}
\label{sec:limitations}

\textbf{Network-level interpretation.} The models operate on whole-brain connectivity without regional priors, and several perform close to chance on some cohorts; Table~\ref{tab:cohorts} and Fig.~\ref{fig:networks} therefore summarise previously reported findings rather than results of this study. Attention weights are available but are not readily interpretable when classification performance is close to chance.

\textbf{Correlation-derived inputs.} Every classifier in this study, quantum or classical, receives features derived from Pearson correlation: selected entries of the correlation vector, its principal components, or regional connectivity profiles. Whether quantum similarity captures non-linear dependence between raw regional BOLD signals, which motivates this work, was therefore not tested; doing so would require encoding the time series directly.

\textbf{Simulation and register width.} The main results come from exact statevector simulation; the kernel result holds under simulated shot and depolarising noise (Section~\ref{sec:robustness}), while gate-level noise and decoherence on hardware remain untested. In the evaluated datasets, the usable fidelity range narrowed as register width increased (Section~\ref{sec:fidelity-collapse}), while feature capacity demands the opposite, and although fidelity concentration need not limit variational models (Section~\ref{sec:width}), simulation cost does. Whether the trend holds on hardware that does not pay that cost remains open.

\textbf{Statistical scope.} The reported intervals are computed across folds rather than across initialisations. Repeating the ABIDE~I suite at four seeds indicates the size of the omitted variation: the largest standard deviation of mean accuracy across seeds is $0.017$ (QSVM-Pegasos), and the two kernel models without a stochastic component are identical at every seed. Differences smaller than $0.017$ between stochastic models should therefore be interpreted cautiously; the matched differences reported here nevertheless keep their sign across seeds, with the single exception of VQC against its matched MLP. The permutation null covers the kernel arm on ABIDE~I only; the neural and graph arms were not permuted. For the graph pipeline, the test is less informative, because every fold of both fidelity-graph models predicts a single class.

\textbf{Graph construction and baselines.} The adjacency is directed and unsymmetrised, and $k$ is fixed at $20$, so the effect of either choice on node distinctiveness and accuracy was not measured. The evaluated baselines were intentionally selected to support matched comparisons rather than to represent the latest adaptive, dynamic or hypergraph learning architectures; re-implementing such a model under leave-site-out folds would position these results relative to the state of the art.

\section{Conclusions}\label{sec-conclusions}
This study benchmarked near-term quantum models against matched classical baselines for functional-connectivity classification on four cohorts, three of them under leave-site-out validation. Most quantum models matched their classical counterparts, but none consistently outperformed them, so the results do not demonstrate a reproducible quantum advantage across the evaluated cohorts.

For fidelity-based graph construction, each added qubit leaves fewer region pairs above every fidelity floor tested, reflecting kernel concentration \cite{Thanasilp2024Concentration} rather than an implementation fault, and any trainable layer applied after the encoding cancels in the fidelity, so trainable similarity must be placed inside or between encodings, or in classical layers after the overlap (Sections~\ref{sec:fidelity-collapse} and~\ref{sec:invariance}). At eight qubits, the quantum kernel and a classical RBF given the same eight features are statistically indistinguishable on all four cohorts across $62$ folds (Section~\ref{sec:matched-comparison}), and under a random-effects model the $95\%$ confidence interval excludes an average advantage above $1.4$ percentage points (Fig.~\ref{fig:bounds}). Four attempts to lift the resulting feature cap, widening the register, learning a projection, ensembling over disjoint blocks, and adding data re-uploading, each helped the matched classical twin at least as much as the quantum arm in the evaluated experiments. Only the parity-read-out architecture consistently underperformed its matched classical counterpart.

One of the most broadly applicable findings concerns model capacity. Models with higher training accuracy did not achieve higher held-out accuracy in the evaluated comparisons: training fit tracks the overfitting gap closely on both well-powered cohorts ($r=+0.95$ on ABIDE~I, $r=+0.98$ on REST-meta-MDD) while relating negatively to held-out accuracy, and capacity added by any mechanism, quantum or classical, raised training fit without raising held-out accuracy. Across $284$ matched comparisons, the quantum arm leads significantly in $30$; $18$ of these involve a predictor that is not classifying or fails to train, and half of the remaining $12$ are on AUC, a \emph{ranking} metric, with none surviving Bonferroni correction across all $284$ comparisons.

These findings are specific to classically simulated circuits, fidelity-based topology and the model families tested here. The matched evaluation protocol and released per-fold results provide a reproducible benchmark for future quantum and classical studies.

\section*{Acknowledgment}
The authors thank the ABIDE, ADHD-200, REST-meta-MDD and UCLA-CNP consortia for collecting and sharing the data analysed here. Fig.~\ref{fig:networks} was rendered with BrainPainter \cite{brainpainter2019}.

\section*{Data Availability and Ethics}
All cohorts were obtained under each consortium's access terms; ethical approval and informed consent were obtained by the contributing sites, and no new data were collected. Code, per-fold results, scripts and fold definitions are available at \url{https://github.com/r11up/qml-fmri-baseline}.

\end{document}